%% file: main.tex
\documentclass[10pt,journal,compsoc]{IEEEtran}

\ifCLASSOPTIONcompsoc
  \usepackage[nocompress]{cite}
\else
  \usepackage{cite}
\fi
\ifCLASSINFOpdf
\else
\fi
\usepackage{amsmath,amsfonts}
\usepackage{algorithmic}
\usepackage{algorithm}
\usepackage{array}
\usepackage{textcomp}
\usepackage{stfloats}
\usepackage{url}
\usepackage{verbatim}
\usepackage{graphicx}
\usepackage{cite}
\newcommand{\figref}[1]{Figure~\ref{#1}}
\newcommand{\tabref}[1]{Table~\ref{#1}}
\newcommand{\secref}[1]{Section~\ref{#1}}

\usepackage{xcolor}
\usepackage{booktabs}
\usepackage{subcaption}
\usepackage{tabularx}
\usepackage[misc]{ifsym}
\usepackage[colorlinks=true, citecolor=blue, linkcolor=red, urlcolor=cyan]{hyperref}

\begin{document}

\title{DreamComposer++: Empowering Diffusion Models with Multi-View Conditions for 3D Content Generation}

\author{
    Yunhan~Yang$^{\ast}$,
    Shuo~Chen$^{\ast}$,
    Yukun~Huang$^{\ast}$,
    Xiaoyang~Wu,
    Yuan-Chen~Guo,~\IEEEmembership{Member,~IEEE,}\\
    Edmund Y. Lam,~\IEEEmembership{Fellow,~IEEE,}
    Hengshuang~Zhao,
    Tong~He,
    Xihui~Liu\textsuperscript{\Letter}~\IEEEmembership{Member,~IEEE}
    \IEEEcompsocitemizethanks{
        \IEEEcompsocthanksitem Y.Yang, Y. Huang, X. Wu, E. Lam, H. Zhao and X. Liu are with The University of Hong Kong (HKU), Hong Kong SAR 999077, China.\\
        E-mail: yhyang07@connect.hku.hk, yukun@hku.hk, \\xiaoyang.wu@connect.hku.hk, elam@eee.hku.hk,
        \\hszhao@cs.hku.hk, xihuiliu@eee.hku.hk
        \IEEEcompsocthanksitem S.Chen is with Tsinghua University, Beijing, China.\\
        E-mail: cs2020012894@gmail.com
        \IEEEcompsocthanksitem Y. Guo is with Vast, Beijing, China.\\
        E-mail: imbennyguo@gmail.com
        \IEEEcompsocthanksitem T. He is with Shanghai AI Laboratory, Shanghai, China.\\
        E-mail: tonghe90@gmail.com
    }
    \thanks{$\ast$: Equal Contribution, \Letter~: Corresponding author.}
}




\IEEEtitleabstractindextext{
\begin{abstract}
\input{tex/0_abstract}
\end{abstract}

\begin{IEEEkeywords}
3D Controllable Generation, Novel View Synthesis, Diffusion Model
\end{IEEEkeywords}
}

\maketitle

\IEEEdisplaynontitleabstractindextext
\IEEEpeerreviewmaketitle

\input{tex/1_introduction}

\input{tex/2_related}

\input{tex/3_method}

\input{tex/4_experiments}

\input{tex/5_conclusion}

\input{tex/acknowledgements}


\bibliographystyle{IEEEtran}
\bibliography{main}



 





\end{document}

%% file: tex/0_abstract.tex
Recent advancements in leveraging pre-trained 2D diffusion models achieve the generation of high-quality novel views from a single in-the-wild image. However, existing works face challenges in producing controllable novel views due to the lack of information from multiple views. In this paper, we present DreamComposer++, a flexible and scalable framework designed to improve current view-aware diffusion models by incorporating multi-view conditions. Specifically, DreamComposer++ utilizes a view-aware 3D lifting module to extract 3D representations of an object from various views. These representations are then aggregated and rendered into the latent features of target view through the multi-view feature fusion module. Finally, the obtained features of target view are integrated into pre-trained image or video diffusion models for novel view synthesis. Experimental results demonstrate that DreamComposer++ seamlessly integrates with cutting-edge view-aware diffusion models and enhances their abilities to generate controllable novel views from multi-view conditions. This advancement facilitates controllable 3D object reconstruction and enables a wide range of applications.

%% file: tex/1_introduction.tex
\IEEEraisesectionheading{\section{Introduction}}
\label{sec:intro}

\begin{figure*}
\centering
\includegraphics[width=\linewidth]{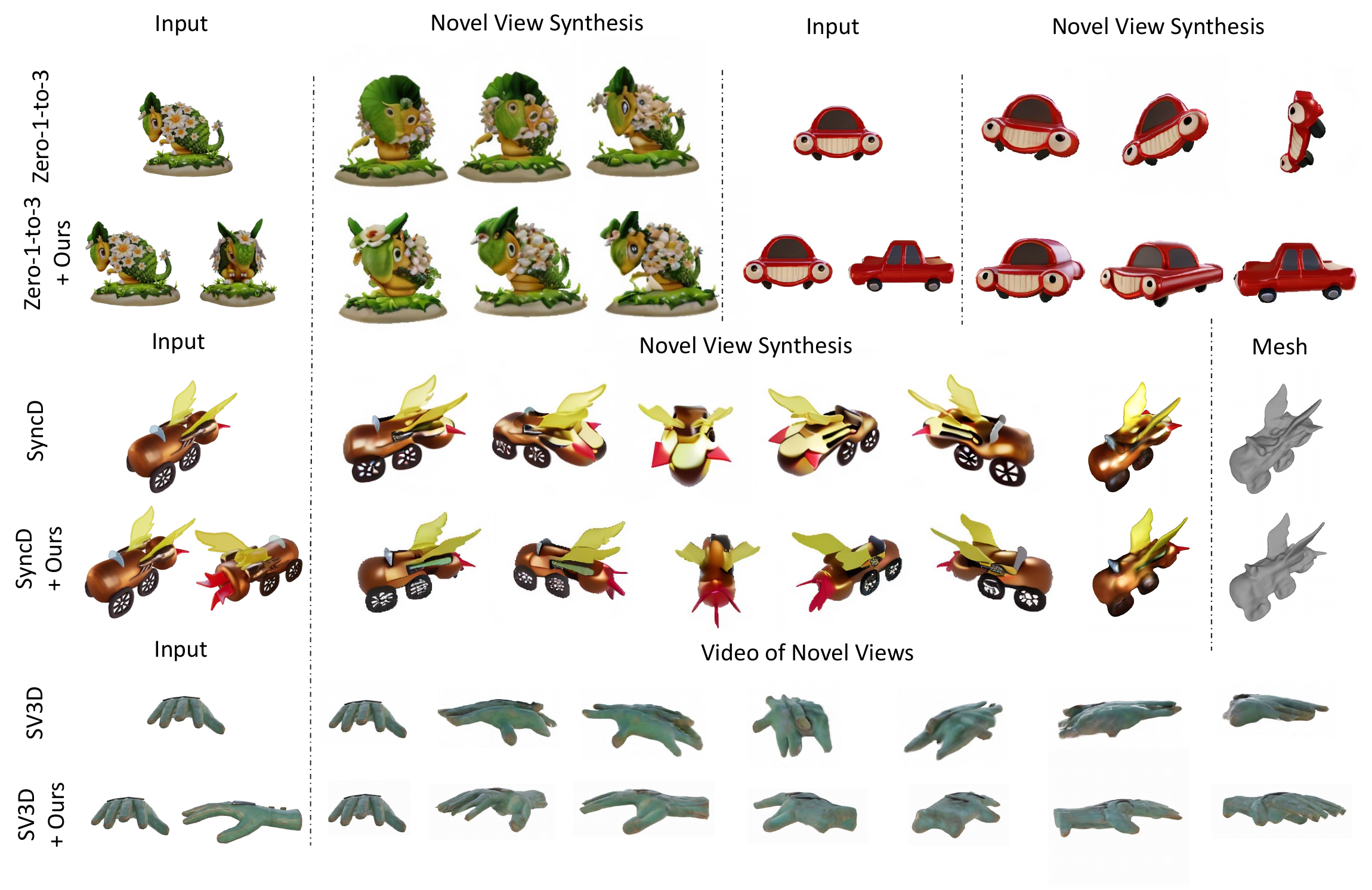}
\captionof{figure}{\textbf{DreamComposer++} is able to generate controllable novel views and 3D objects via injecting multi-view conditions. We incorporate the method into the pipelines of Zero-1-to-3~\cite{liu2023zero1to3}, SyncDreamer (SyncD)~\cite{liu2023syncdreamer} and SV3D~\cite{voleti2024sv3d} to enhance the control ability of those models.}
\label{fig:teaser}
\end{figure*}

\IEEEPARstart{T}{he} generation of 3D objects has become a prominent area of study in the fields of computer vision and graphics. This research area has significant implications for various applications, including augmented reality (AR), virtual reality (VR), film production, and the gaming industry. Leveraging 3D object generative models, users and designers are empowered to efficiently produce customized 3D assets using textual or visual cues, minimizing the need for extensive manual labor by human experts.

Recently, diffusion models~\cite{ho2020denoising, rombach2021highresolution} have achieved significant success in generating 2D images from textual descriptions. This breakthrough has sparked interest in extending these models to 3D object generation, leveraging \textit{2D diffusion priors}~\cite{poole2022dreamfusion, wang2023score, huang2023dreamtime, wang2023prolificdreamer, lin2023magic3d, chen2023fantasia3d} as supervision to learn 3D representations via \textit{SDS loss}. While impressive results have been achieved~\cite{wang2023prolificdreamer}, 2D diffusion models suffer from limited view control and lack the ability to ensure view-consistent supervision. This often leads to quality issues in 3D outputs, such as the appearance of multiple faces and blurry details. To address this issue, Zero-1-to-3~\cite{liu2023zero1to3} finetunes 2D diffusion models with \textit{view conditioning}, enabling zero-shot novel view synthesis (NVS) based on a single-view image. This approach facilitates image-to-3D object generation, improving consistency across different views. Follow-up works~\cite{liu2023syncdreamer, shi2023zero123plus, ye2023consistent1to3, weng2023consistent123, long2023wonder3d} further propose to enhance the \textit{cross-view consistency} of generated images. However, due to the incomplete information from a single-view input, these methods often struggle with generating unpredictable or implausible shapes and textures when synthesizing novel views. For example, by only observing the side view of the shoes, it is impossible to determine whether there is one shoe or a pair. In other words, novel view synthesis and 3D object generation remain difficult to generate reliably when conditioned solely on a single-view image. 
Despite efforts to improve synthesis fidelity, current methods~\cite{zou2023sparse3d,yu2021pixelnerf,zhou2023sparsefusion,reizenstein2021common} still struggle with producing realistic 3D shapes from sparse or ambiguous inputs, particularly in complex and unconstrained real-world settings. This limitation highlights the need for more robust approaches capable of handling diverse, in-the-wild data distributions and capturing fine-grained 3D structure under limited observations.

In this work, we aim to introduce flexible multi-view image conditioning to diffusion models, enabling more controllable novel view synthesis and 3D object generation for real-world objects. For instance, given front, back, and side views of an object provided by designers, the model can generate novel views while maintaining consistency with the input views. It also enables interactive 3D generation, allowing users to provide additional views if the generated 3D objects deviate from their intended design. However, this task poses two major challenges. First, integrating an arbitrary number of input views into a coherent signal that can guide the generation of novel views is non-trivial. Secondly, it is tricky to ensure compatibility across different network architectures, such as Zero-1-to-3~\cite{liu2023zero1to3}, SyncDreamer~\cite{liu2023syncdreamer}, and SV3D~\cite{voleti2024sv3d}.

In this work, we introduce DreamComposer++, a scalable and flexible framework that extends existing view-aware diffusion models to accommodate an arbitrary number of multi-view inputs. DreamComposer++ consists of three key stages: target-aware 3D lifting, multi-view feature fusion, and target-view feature injection.
(i) Target-Aware 3D Lifting encodes multi-view images into latent space and subsequently lifts these latent features to 3D tri-planes~\cite{chan2022efficient}. This tri-plane representation, enhanced with latent features, is both compact and efficient. The target-view-aware design enables the network to concentrate on constructing 3D features pertinent to the target view.
(ii) Multi-View Feature Fusion renders and fuses the 3D features from different views into target-view 2D features using a novel composited volume rendering approach.
(iii) Target-View Feature Injection integrates the latent features from the previous stage into the diffusion models using a ControlNet-like structure. This injection module utilizes the relative angle as a condition, allowing for adaptive gating of the multi-view inputs.
DreamComposer++ can be seamlessly integrated into existing image and video diffusion models, such as Zero-1-to-3~\cite{liu2023zero1to3}, SyncDreamer~\cite{liu2023syncdreamer}, and SV3D~\cite{voleti2024sv3d}, enhancing their capabilities with multi-view conditioning, as illustrated in \figref{fig:teaser}.

Our contributions are summarized as follows:
\begin{itemize}
\item We propose DreamComposer++, a flexible multi-view conditioning framework that enhances existing image and video diffusion models for scalable novel view synthesis.
\item We design three key modules: target-aware 3D lifting, multi-view feature fusion, and target-view feature injection. These modules underpin the scalability and flexibility of DreamComposer++.
\item Experiments demonstrate that our framework is compatible with existing state-of-the-art view-aware diffusion models, achieving high-fidelity novel view synthesis and controllable 3D object generation.
\item Benefiting from multi-view conditioning, our framework further supports a wide range of applications, including 3D object editing and character modeling.
\end{itemize}

Compared with the preliminary conference version~\cite{yang2024dreamcomposer}, DreamComposer++ introduces several non-trivial improvements. The most notable improvement is the support for video diffusion model, which enables more dense and consistent multi-view generation compared to previous image diffusion-based models. To better adapt to video diffusion models, we introduce two key modules: \textit{View-aware Attention}, which preserves the consistency of multi-view latent representations, and \textit{Shifted Window Cross Attention}, which enhances model robustness. By empowering SV3D~\cite{voleti2024sv3d} with multi-view conditioning, we demonstrate further improvements in both generation quality and controllability. In addition, we conduct more ablation analysis and visualizations to highlight the effectiveness of DreamComposer++. Finally, DreamComposer++ achieves higher resolution and finer texture control, making it suitable for more complex applications than the previous version~\cite{yang2024dreamcomposer}.

%% file: tex/2_related.tex
\section{Related Work}
\label{sec:related}

\noindent\textbf{Zero-shot Novel View Synthesis.} Previous works~\cite{mildenhall2021nerf,kulhanek2022viewformer,gu2023nerfdiff} on novel view synthesis are generally trained on datasets with limited scenes or categories and cannot generalize to in-the-wild image inputs.
Recently, diffusion models~\cite{rombach2021highresolution,saharia2022imagen} trained on large-scale Internet data have demonstrated powerful open-domain text-to-image generation capabilities. This success inspired the community to implement zero-shot novel view synthesis by fine-tuning these pre-trained diffusion models. Zero-1-to-3~\cite{liu2023zero1to3} fine-tuned the Stable Diffusion model~\cite{rombach2021highresolution} on the large 3D dataset Objaverse~\cite{deitke2022objaverse}, achieving viewpoint-conditioned image synthesis of an object from a single in-the-wild image.
Based on Zero-1-to-3, several subsequent works~\cite{liu2023syncdreamer, shi2023zero123plus, ye2023consistent1to3, weng2023consistent123, long2023wonder3d, liu2023one2345, xu2024instantmesh, wang2024crm, liu2023one2345++, gao2024cat3d, zheng2023free3D, wang2023imagedream, shi2023toss} aim to produce multi-view consistent images from a single input image to create high-quality 3D objects. However, limited by the ambiguous information of single input image, these models might produce uncontrollable results when rendering novel views.

\noindent\textbf{Diffusion Models with 3D Priors.} In addition to fine-tuning directly on the pre-trained text-image diffusion models,
some recent works~\cite{zou2023sparse3d, zhou2023sparsefusion, kulhanek2022viewformer, watson2022novel, chan2023generative, gu2023nerfdiff, wen2024ouroboros3d, huang2024dreamcontrol, sweetdreamer, lin2024dreampolisher} also attempt to combine diffusion models with 3D priors for novel view synthesis. GeNVS~\cite{chan2023generative} integrates geometry priors in the form of a 3D feature volume into the 2D diffusion backbone, producing high-quality, multi-view-consistent renderings on varied datasets. NerfDiff~\cite{gu2023nerfdiff} distills the knowledge of a 3D-aware conditional diffusion model into NeRF at test-time, avoiding blurry renderings caused by severe occlusion. While remarkable outcomes have been obtained for particular object categories from ShapeNet~\cite{chang2015shapenet} or Co3D~\cite{reizenstein2021common}, the challenge of designing a generalizable 3D-aware diffusion model for novel view synthesis from any in-the-wild inputs remains unresolved.

\noindent\textbf{Video Diffusion Models for Novel View Synthesis.} Video diffusion models incorporate temporal modules to maintain frame consistency, making them particularly effective for generating coherent multi-view outputs. Some recent works~\cite{voleti2024sv3d, melas20243d, zuo2024videomv, yu2024viewcrafter, yang2024hi3d, you2024nvs} fine-tune pre-trained image-to-video diffusion models to create novel views of 3D contents. Leveraging the strengths of video diffusion models, these approaches achieve more consistent multi-angle views, leading to enhanced 3D reconstruction quality. However, since they only accept a single view as input, the controllability of video generation is limited.

\noindent\textbf{3D Generation with 2D Diffusion Models.} Due to the limited size of existing 3D datasets, it remains challenging to train generative 3D diffusion models~\cite{jun2023shape, nichol2022pointe, wang2022rodin, müller2023diffrf} using 3D data. With pre-trained text-to-image diffusion models and score distillation sampling~\cite{poole2022dreamfusion}, DreamFusion-like methods~\cite{poole2022dreamfusion, wang2023score, wang2023prolificdreamer, lin2023magic3d, chen2023fantasia3d, metzer2022latentnerf, huang2023dreamtime, yi2023gaussiandreamer, wu2024consistent3d, alldieck2024score, tang2023stable, yan2024DreamView} have achieved remarkable text-to-3D object generation by distilling 2D image priors into 3D representations. Some methods~\cite{melas2023realfusion, tang2023make, xu2023neurallift, qian2023magic123, tang2023makeit3d, xu2023neurallift360, sun2023dreamcraft3d} utilize similar distillation approaches to execute image-to-3D tasks. Since these works, which rely on an optimization strategy, have not previously encountered real 3D datasets, they face the Janus (multi-face) problem, making it challenging to generate high-quality 3D object shapes.

\noindent\textbf{Conditional 3D Object Generation.} In addition to generating 3D content conditioned on text or a single image, recent studies have explored multi-view or 3D conditioning to improve controllability.
Building upon MeshGPT~\cite{siddiqui2024meshgpt}, several recent approaches~\cite{chen2024meshanything, chen2024meshanythingv2, chen2024meshxl, weng2024scaling, tang2024edgerunner, hao2024meshtron} adopt 3D point clouds as conditioning inputs and utilize autoregressive mechanisms to directly generate mesh faces. These methods yield meshes more closely aligned with artist-created quality. 
Trellis~\cite{xiang2024structured} employs a two-stage generation mechanism. In the first stage, a coarse shape voxel representation is generated conditioned on text or an image. Subsequently, in the second stage, it produces a refined mesh conditioned on the initial coarse voxel shape.
The Large View Synthesis Model (LVSM)~\cite{jin2024lvsm} proposes a transformer-based framework for scalable and generalizable novel view synthesis from sparse-view inputs. The input views and corresponding camera embeddings are tokenized and fed into transformer blocks, enabling the generation of the target view conditioned on its camera embedding.
Unlike LVSM, which lacks explicit 3D modeling, our approach enhances flexibility for multi-view inputs through explicit 3D spatial modeling.
Our framework is compatible with any pre-trained diffusion model without requiring fine-tuning them, thereby preserving the original performance of the base model. By incorporating our proposed method, various diffusion models can be readily augmented with multi-view input controllability.

%% file: tex/3_method.tex
\begin{figure*}
\centering
\includegraphics[width=\linewidth]{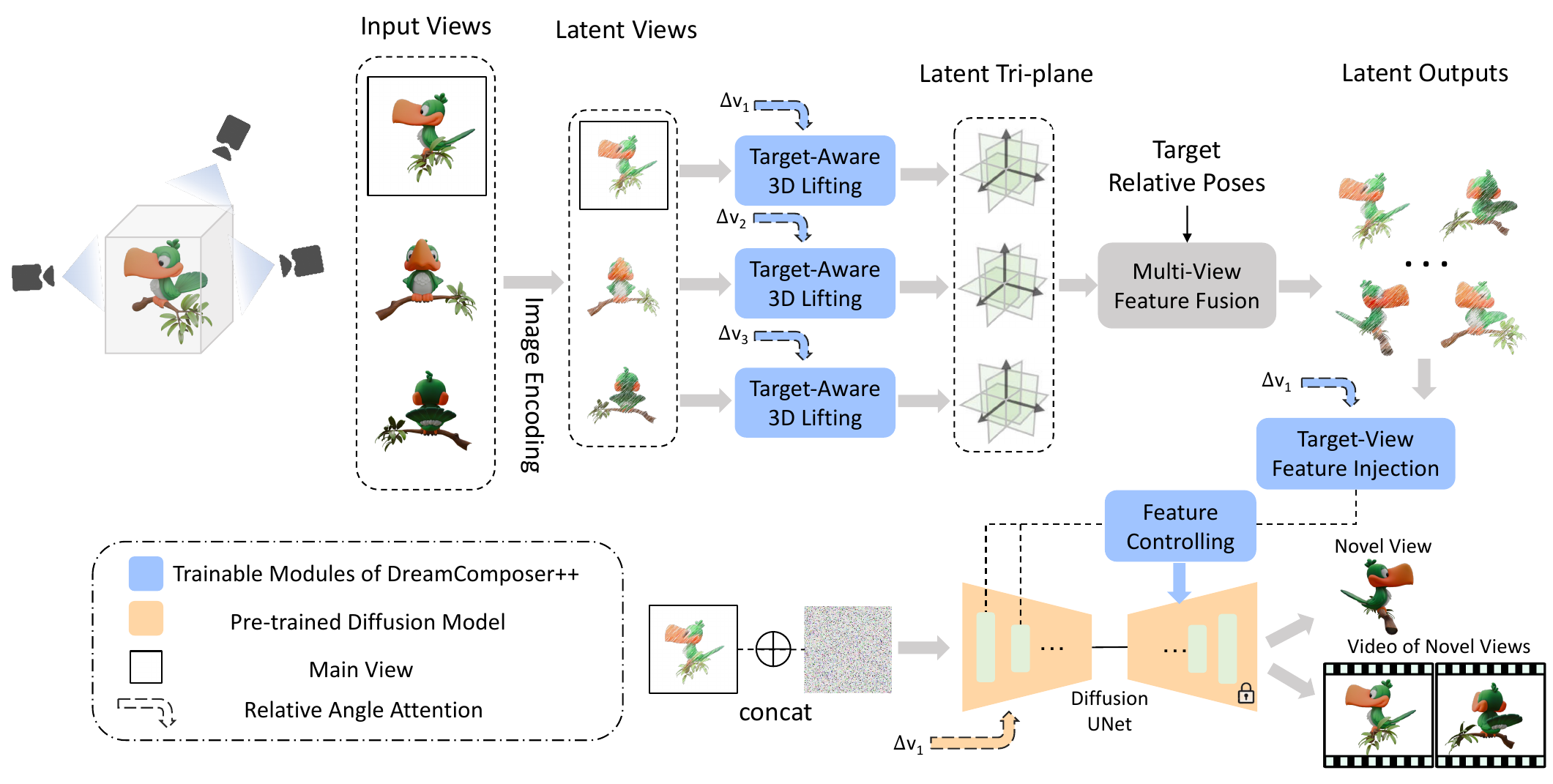}
\caption{An overview pipeline of \textbf{DreamComposer++}. Given multiple input images from different views, DreamComposer++ extracts their 2D latent features and uses a 3D lifting module to produce tri-plane 3D representations. Then, the multi-view condition rendered from 3D representations is injected into the pre-trained diffusion model to provide target-view auxiliary information.}
\label{fig:pipeline}
\end{figure*}

\begin{figure}
    \centering
   \includegraphics[width=1.0\linewidth]{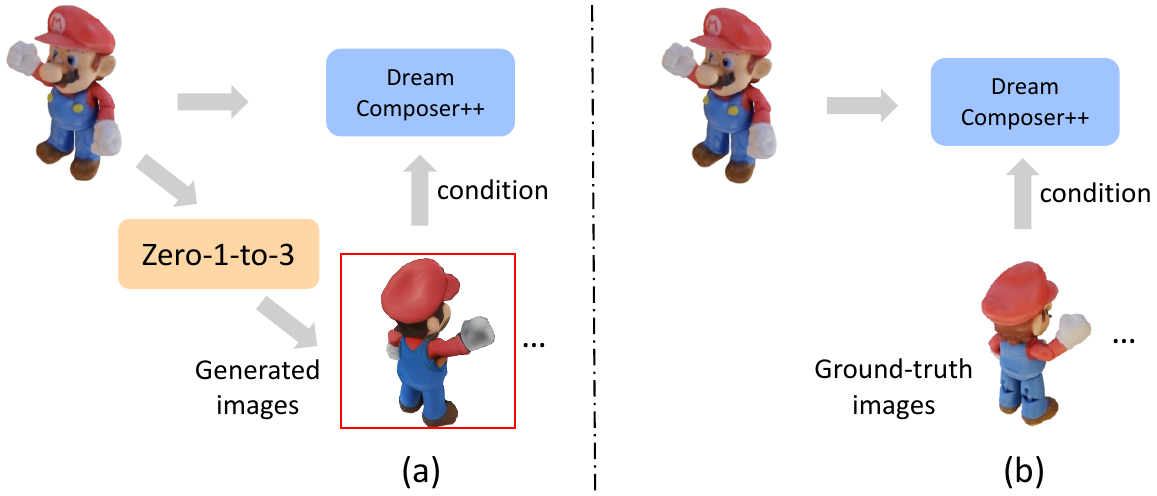}
    \caption{\textbf{Different numbers of ground-truth inputs.} Our model is capable of handling a variety of ground-truth input quantities.}
   \label{fig:input-num-vis}
\end{figure}

\section{Method}
\label{sec:method}
DreamComposer++ aims to enhance the controllability and flexibility of diffusion models conditioned on multi-view inputs.
To obtain latent representations of target views from arbitrary input views, we propose the \textbf{Target-aware 3D Lifting} module, which elevates 2D features from the input views into a unified 3D latent representation (detailed in Sec.\ref{sec:3d_lifting}).
Subsequently, to obtain the target views' latent outputs from all inputs, we introduce a \textbf{Multi-View Feature Fusion} module (described in Sec.\ref{sec:view_fusion}), which employs a novel composited volume rendering approach to render and fuse the 3D latent features from multiple input views into the 2D latent features of the target views.
Finally, the obtained target-view latent outputs are integrated into the diffusion models through our \textbf{Target-View Feature Injection} module (Sec.~\ref{sec:injection}), enabling controllable multi-view novel view synthesis.

\subsection{Preliminaries: Multi-View Diffusion}

Most multi-view diffusion models~\cite{liu2023zero1to3} take a single posed image as input and then generate target view images conditioned on specific target poses. Given a single RGB image \( I_x \in \mathbb{R}^{H \times W \times 3} \), multi-view diffusion, represented as \( f_\theta \), can generate multi-view images \( \hat{I_i} \) based on the target camera rotation \( R_i \) and translation \( T_i \):

\begin{equation}
    \hat{I_i} = f_\theta(I_x, R_i, T_i)
\end{equation}

Here, \( f_\theta \) represents the diffusion process, which iteratively denoises and refines the output images conditioned on the given pose parameters. The goal is to learn a mapping from the source view to various target views through this iterative denoising mechanism.

To enhance the control and fidelity of the generated multi-view images, we propose incorporating multiple view inputs during the diffusion process. This allows the model to generate more consistent and controllable multi-view outputs by leveraging additional pose-aligned image data. Formally, given a reference image \( I_x \) and a set of additional posed images \( \{ I_{a_1}, I_{a_2}, \dots, I_{a_n} \} \), we extend the diffusion model to take these additional views into account when generating the target image \( \hat{I_i} \):

\begin{equation}
\begin{split}
    \hat{I_i} = f_\theta(I_x, \{ I_{a_1}, I_{a_2}, \dots, I_{a_n} \}, R_x, T_x, \\
    \hspace{22mm} \{ R_{a_1}, T_{a_1}, \dots, R_{a_n}, T_{a_n} \})
\end{split}
\end{equation}

In this formulation, the additional posed images \( \{ I_{a_1}, I_{a_2}, \dots, I_{a_n} \} \) serve as auxiliary inputs that provide extra information about the scene geometry and texture from multiple viewpoints. By conditioning on these additional views, the diffusion model can generate more accurate and consistent results, especially when reconstructing occluded or ambiguous regions in the target view.

The diffusion process itself is governed by the following stochastic differential equation (SDE):

\begin{equation}
    d\mathbf{x}_t = \mathbf{f}(\mathbf{x}_t, t) dt + g(t) d\mathbf{w}_t
\end{equation}

Here, \( \mathbf{x}_t \) represents the noisy data at time step \( t \), \( \mathbf{f} \) is the drift function, and \( g(t) \) controls the diffusion magnitude. The addition of the multi-view inputs modifies the drift function, allowing the model to guide the diffusion process towards the desired target views with more precision and consistency. This approach significantly improves the accuracy and coherence of the generated multi-view images.

\subsection{Target-Aware 3D Lifting}\label{sec:3d_lifting}

Traditional diffusion models~\cite{liu2023zero1to3,liu2023syncdreamer} focus primarily on single-view input, limiting their capacity to process an undefined number of multi-view inputs effectively. To overcome this, we propose a method to project 2D features from different viewpoints into a unified 3D space, facilitating view-conditioned transformations in a scalable and efficient manner.

\noindent\textbf{2D-to-3D Feature Projection.} For an input image $\mathbf{I}_i \in \mathbb{R}^{H \times W \times C}$ captured from camera view $i$, we first encode the image using a modified image encoder $\mathcal{E}$, which maps the input to a latent space feature $\mathbf{z}_i \in \mathbb{R}^{H' \times W' \times d}$. Here, $H'$ and $W'$ represent the spatial resolution in the latent space, and $d$ denotes the feature dimensionality. 

Next, we introduce a 3D projection module, parameterized by $\mathbf{P}$, that takes these latent 2D features and lifts them into 3D feature space. This module relies on both convolutional operations and attention mechanisms to project the 2D latent feature $\mathbf{z}_i$ into a 3D tensor $\mathbf{Z}_i \in \mathbb{R}^{H' \times W' \times D \times 3}$, where $D$ is the depth in 3D space. This projection process is conditioned on the relative rotation $\Delta \alpha_i$, the angular difference between the current input view and a target view.

We employ a tri-coordinate system $\mathbf{Z}_i = \{\mathbf{Z}_i^{xy}, \mathbf{Z}_i^{xz}, \mathbf{Z}_i^{yz}\}$~\cite{chan2022efficient}, as it balances computational efficiency and representation capacity. The projection happens directly in latent space, minimizing the computational overhead by avoiding explicit 3D reconstructions at pixel level.

\noindent\textbf{View-conditioned Attention Mechanism.} The 3D projection module $\mathbf{P}$ includes self-attention layers that process the latent features within each view and cross-attention layers to handle interactions between input and target views. By introducing the angular difference $\Delta \alpha_i$ as a conditioning factor, we ensure that the 3D projection focuses on building features most relevant to the desired target view, instead of constructing a full 3D scene from each input.

Formally, the output 3D representation for each view is given by:
\begin{equation}
    \mathbf{Z}_i = \mathbf{P}(\mathbf{z}_i, \Delta \alpha_i)
\end{equation}

This ensures adaptive feature projection, making the 3D representation sensitive to the viewing direction.

\noindent\textbf{Handling Multi-View Inputs.} When we are given images from multiple camera views, say $\mathbf{I}_1, \mathbf{I}_2, \dots, \mathbf{I}_n$, we process each view independently through the image encoder and project each into its corresponding 3D feature representation $\mathbf{Z}_1, \mathbf{Z}_2, \dots, \mathbf{Z}_n$. These multi-view 3D features are then available for further processing to synthesize new views or perform view-conditioned transformations.

The overall process for multi-view feature lifting can be represented as:
\begin{equation}
    \mathbf{Z}_{1:n} = \{\mathbf{P}(\mathbf{z}_i, \Delta \alpha_i)\}_{i=1}^{n}
\end{equation}

These projected 3D features from different views serve as auxiliary inputs in subsequent stages of the diffusion process, providing a robust, view-aware understanding of the scene.

\subsection{Multi-View Feature Fusion}\label{sec:view_fusion}

After extracting 3D features $\{\mathbf{Z}_1, \mathbf{Z}_2, \dots, \mathbf{Z}_n\}$ from the $n$ input views, our goal is to generate a target-view latent feature $\mathbf{z}_t$ based on these 3D features, which will serve as the conditioning input for the diffusion model.

\noindent\textbf{3D Feature Fusion.} Since each input view represents different perspectives of the scene, fusing these 3D features is non-trivial due to the misalignment between the camera spaces of the different views. To address this, we employ a volume rendering-based approach for 3D feature fusion. This process involves the following steps: (1) sampling ray points from the target view, (2) transforming these points to the coordinate systems of the input views, (3) retrieving and aggregating 3D features from each input view, and (4) integrating the fused features along the target-view rays to form the target-view latent feature $\mathbf{z}_t$.

Given a sampled 3D point $\mathbf{p}_t$ along a ray from the target view, we project this point into the camera space of the $i$-th input view using the transformation matrix $\mathbf{T}_{t \to i}$:

\begin{equation}
    \mathbf{p}_i = \mathbf{T}_{t \to i} \mathbf{p}_t
\end{equation}

We then retrieve the corresponding feature $\mathbf{f}_p^i$ from the input view $i$ at the location $\mathbf{p}_i$. This process is repeated for all input views, resulting in a set of 3D point features $\{\mathbf{f}_p^1, \mathbf{f}_p^2, \dots, \mathbf{f}_p^n\}$ for the sampled point $\mathbf{p}_t$.

\noindent\textbf{Adaptive Weighting for Fusion.} Since the contribution of each input view to the target view depends on the relative viewing angles, we apply an adaptive weighting strategy to combine the features from different views. The azimuthal angle differences between each input view and the target view are denoted as $\Delta \gamma_1, \Delta \gamma_2, \dots, \Delta \gamma_n$. The weight for each input view is computed as follows:

\begin{equation}
    \lambda_i = \frac{\cos(\Delta \gamma_i) + 1}{2}
\end{equation}

These weights are then normalized so that they sum to 1, resulting in the normalized weights $\hat{\lambda}_i$:

\begin{equation}
    \hat{\lambda}_i = \frac{\lambda_i}{\sum_{j=1}^{n} \lambda_j}
\end{equation}

The fused feature for point $\mathbf{p}_t$ is obtained by taking a weighted sum of the features from all input views:

\begin{equation}
    \mathbf{f}_p^t = \sum_{i=1}^n \hat{\lambda}_i \cdot \mathbf{f}_p^i
\end{equation}

\noindent\textbf{Rendering the Target-View Latent Feature.} After calculating the fused features for all sampled points along the rays in the target view, we integrate these features using a volume rendering method, following the approach outlined in~\cite{mildenhall2021nerf}. This integration gives us the target-view latent feature $\mathbf{z}_t$, which is used as the condition for subsequent processing:

\begin{equation}
    \mathbf{z}_t = \int \mathbf{f}_p^t \, d\mathbf{p}_t
\end{equation}

By fusing multi-view 3D features in this adaptive manner, we create a coherent target-view representation that is both geometry-aware and view-consistent, enabling high-quality target-view generation.

\subsection{Target-View Feature Injection}\label{sec:injection} 

The latent feature $\mathbf{f}_t$ encodes rich information from the target view, extracted from multi-view inputs. To leverage this feature for multi-view conditioning, we inject $\mathbf{f}_t$ into the UNet of the diffusion model to guide the generation process. In contrast to the residual-based injection method, we adopt a Concat-conv approach. Specifically, the target-view feature $\mathbf{f}_t$ is concatenated with the intermediate outputs of the UNet and passed through convolutional layers to adaptively influence the generation.

\begin{figure*}
    \centering
    \includegraphics[width=0.95\linewidth]{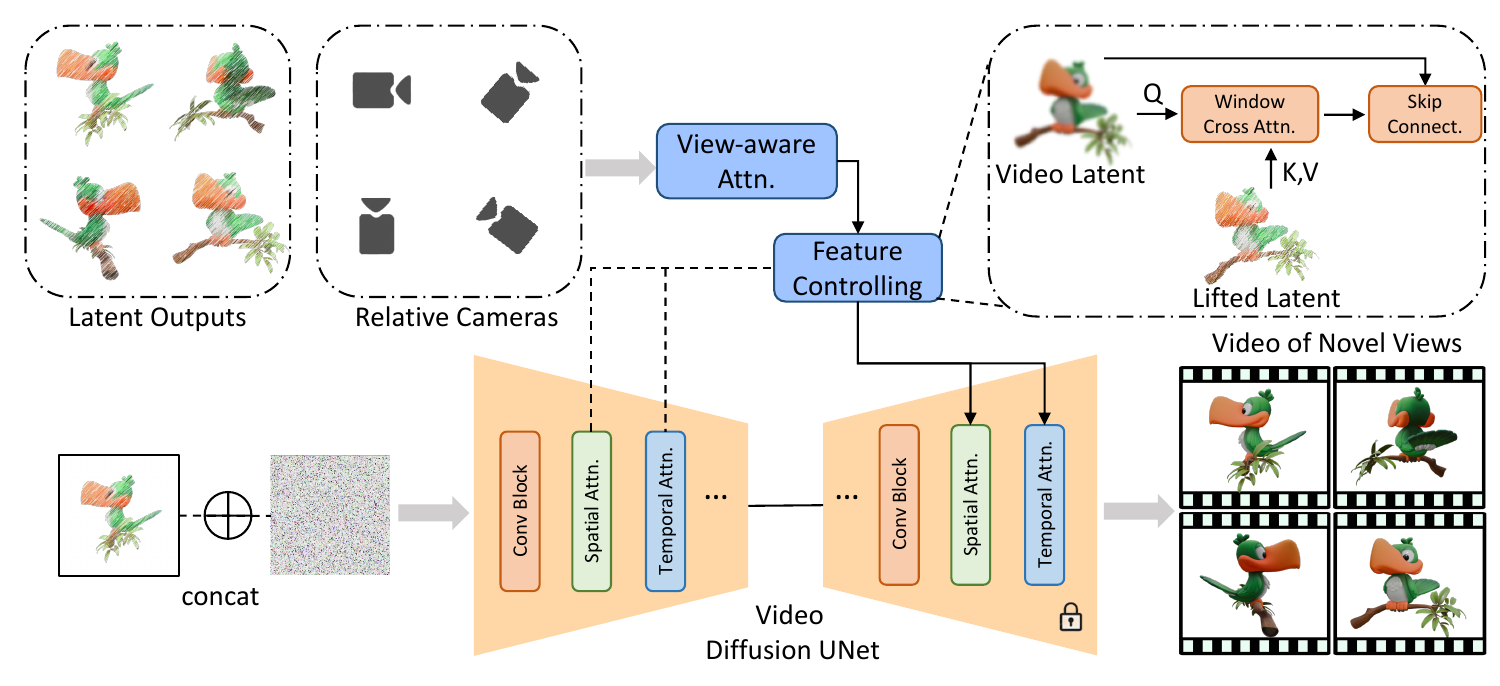}
   \caption{The design of View-aware Attention and Shifted Window Cross-attention for integration with video diffusion models. We employ View-aware Attention to maintain the consistency of multi-view latent information, and Shifted Window Cross Attention to enhance the model's robustness.}
   \label{fig:pipeline_sv3d}
\end{figure*}

\noindent\textbf{Feature Controlling Module.} At each block of the UNet, the target-view feature $\mathbf{f}_t$ is concatenated with the intermediate output of the block to enrich the feature representation. Let $\mathbf{o}_k$ be the intermediate output of the $k$-th block in the UNet. The output after injecting $\mathbf{f}_t$ using the Concat-conv method is defined as:

\begin{equation}
    \mathbf{o}_k' = \text{Conv}([\mathbf{o}_k, \mathbf{f}_t]) + \mathbf{o}_k
\end{equation}

where $[\mathbf{o}_k, \mathbf{f}_t]$ denotes the concatenation of the intermediate output $\mathbf{o}_k$ and the target-view latent feature $\mathbf{f}_t$. The output is then passed through a convolutional layer, followed by the addition of the original output $\mathbf{o}_k$ in a residual connection. This convolutional layer is learned during training and adjusts the combined features to match the target-view conditions.

\noindent\textbf{Comparison with Residual Injection.} Unlike the residual-based method introduced in DreamComposer, where the target-view feature $\mathbf{f}_t$ generates residuals that are added to the intermediate output, the Concat-conv approach directly merges $\mathbf{f}_t$ with $\mathbf{o}_k$ and then passes the concatenated result through a learnable convolutional layer before adding it back to the original output. This combination of concatenation and residual connection allows for richer interactions between the feature sets, enabling more expressive feature fusion. Additionally, the Concat-conv method provides a more direct and controllable influence over the feature representation, as the convolution operation better adjusts both spatial and channel-wise information for multi-view conditioning, while the residual connection ensures the preservation of original features, enhancing the model's ability to handle complex target-view conditions.

\subsection{Training and Inference}\label{sec:multi_view_training}

In the previous sections, we introduced three main components: target-aware 3D lifting, multi-view feature fusion, and target-view feature injection. These components empower the diffusion model with multi-view conditioning capabilities. Among these, the 3D lifting and target-view injection modules are trainable. We propose a two-stage training paradigm to optimize these modules efficiently.

\noindent\textbf{Stage 1: Pre-training 3D Lifting.} In the first stage, we pre-train the target-aware 3D lifting module on a proxy task: sparse view reconstruction. Given several input views, the 3D lifting module is trained to predict novel views using a mean square error (MSE) loss in the latent space:

\begin{equation}
    \mathcal{L}_{\text{lift}} = \frac{1}{M} \sum_{i=1}^{M} \left\| \mathbf{z}_i - \hat{\mathbf{z}}_i \right\|_2^2
\end{equation}

where $\mathbf{z}_i$ is the ground truth latent feature of the novel view, and $\hat{\mathbf{z}}_i$ is the predicted latent feature. This trains the 3D lifting module to generate accurate multi-view representations.

\noindent\textbf{Stage 2: Multi-View Conditioning.} In the second stage, the pre-trained 3D lifting module and a pre-trained diffusion model such as Zero-1-to-3~\cite{liu2023zero1to3} are integrated. We optimize the entire system jointly, including the multi-view feature fusion and target-view feature injection modules. The overall loss function consists of the diffusion loss and the MSE loss from Stage 1:

\begin{equation}
    \mathcal{L} = \mathcal{L}_{\text{diffusion}} + \lambda_{\text{lift}}\mathcal{L}_{\text{lift}}
\end{equation}

where $\lambda_{\text{lift}}$ is a balancing factor that controls the contribution of the 3D lifting loss.

\noindent\textbf{Inference.} During inference, the trained model can accept one or more input views and generate novel views under multi-view conditions. The multi-view conditioning mechanism improves the model’s scalability and controllability, allowing it to generalize well to novel view synthesis and related 3D tasks.

\subsection{DreamComposer++ with Video Diffusion Model}\label{sec:video_diffusion}

Different from DreamComposer~\cite{yang2024dreamcomposer}, we extend our framework to support video generation by integrating the target-aware 3D lifting module with a video diffusion model. This enhanced version, DreamComposer++, enables the generation of dense, high-resolution, and temporally consistent videos, while providing flexible control over different resolutions and viewpoints, as illustrated in Figure \ref{fig:pipeline}.

\noindent\textbf{Flexible 3D Lifting for Video.} The 3D lifting module in DreamComposer++ is highly adaptable, capable of outputting 3D features that can be adjusted to match various resolutions and camera angles. This adaptability is crucial for video diffusion models, as it ensures that the lifted 3D features fit the requirements for video synthesis or multi-view image generation. By adjusting the resolution and perspectives of the 3D lifted features, DreamComposer++ is scalable across both video and image diffusion tasks.

Given a sequence of video frames, the 3D lifting module generates corresponding latent features conditioned on both the temporal evolution of the video and the desired viewpoints. This ensures the system can handle different resolutions and angles, enabling high-quality video outputs.

\noindent\textbf{View-aware Attention.} The spatial and temporal attention modules within the video diffusion model ensure consistency in the generated videos. To better integrate the latent outputs of target views into the video diffusion model and enhance controllability while preserving consistency, we introduce a view-aware attention mechanism to process the latent representations from the 3D Lifting module. Specifically, as shown in \figref{fig:pipeline_sv3d}, we encode these latent outputs together with their corresponding relative camera positions and then feed them into the view-aware attention module. The output features are used for the following feature controlling module.

\noindent\textbf{Shifted Window Cross Attention.} Sometimes, slight deflections or translations may occur in the lifted latent representations generated by the 3D Lifting module, causing misalignment at the pixel level with the video latent features in the diffusion UNet. To enhance robustness against such misalignments, we employ a shifted-window cross-attention mechanism within the feature controlling module. Specifically, as shown in \figref{fig:pipeline_sv3d}, the video latent acts as the query (Q) to extract relevant features from the lifted latent serving as the keys and values (K, V). Additionally, a skip connection is introduced to facilitate direct feature transfer from the video latent.

\noindent\textbf{Video Diffusion with Multi-View Control.} DreamComposer++ allows video diffusion models to utilize multi-view inputs by conditioning the generation process on the 3D lifted features. These features provide essential context, helping the model maintain spatial and temporal consistency across frames, even when the viewpoints change. As a result, DreamComposer++ produces videos that are smooth in time and coherent across different perspectives, offering enhanced resolution and visual fidelity for complex objects, as shown in Figure \ref{fig:sv3d_nvs}.

The combination of flexible 3D lifting and multi-view conditioning ensures that DreamComposer++ generates high-quality, scalable videos with strong control over resolution and viewpoints, making it suitable for various video synthesis tasks.

\begin{figure}
    \centering
    \includegraphics[width=0.8\linewidth]{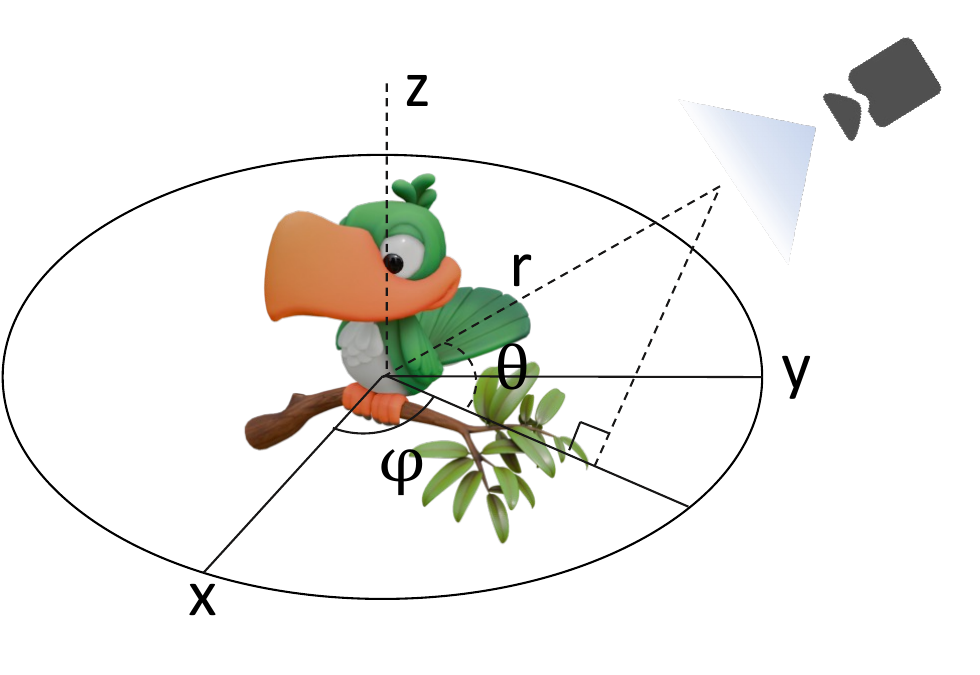}
    \caption{Spherical Coordinate System.}
    \label{fig:cam_system}
\end{figure}

\begin{figure*}
    \centering
    \includegraphics[width=1\linewidth]{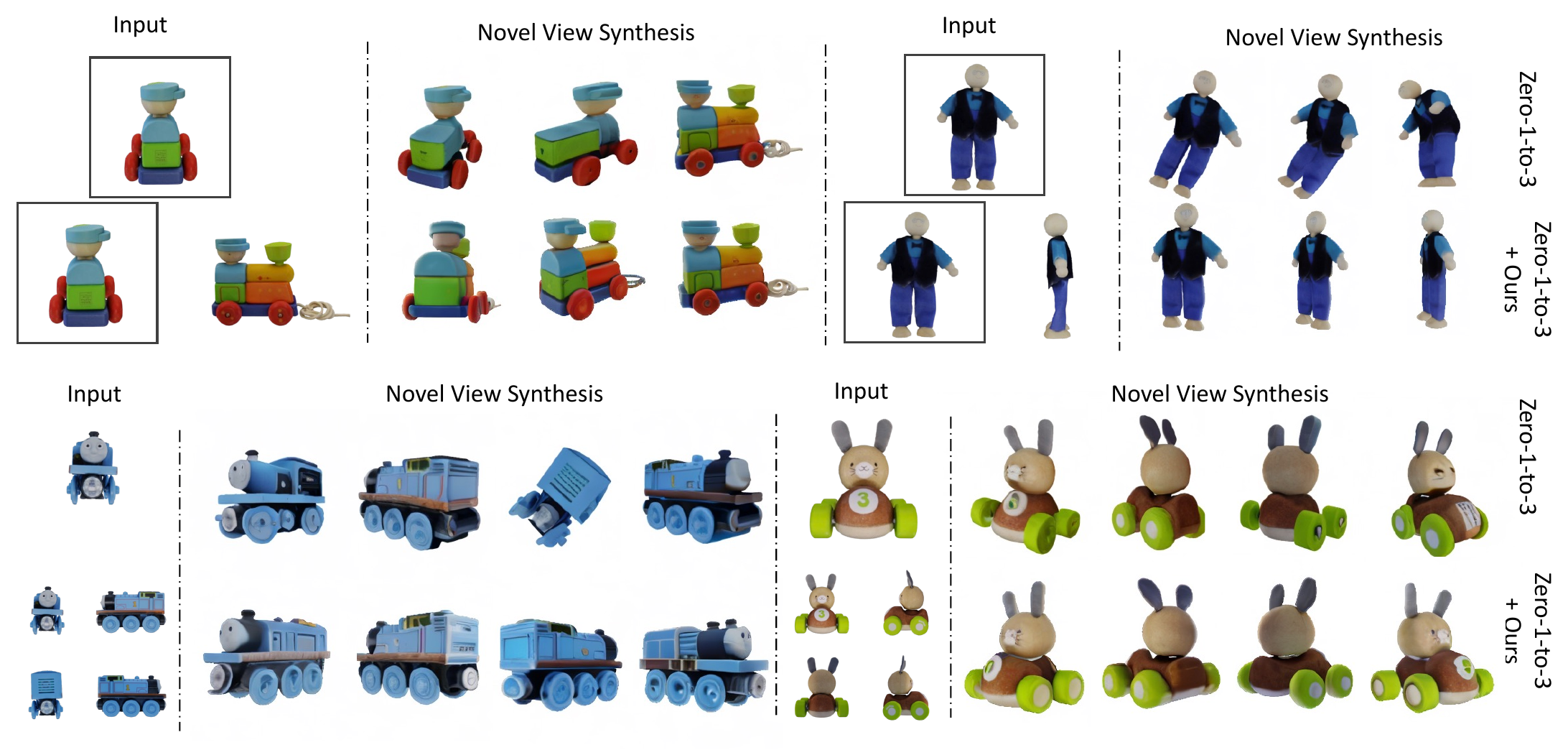}
   \caption{Qualitative comparisons with Zero-1-to-3~\cite{liu2023zero1to3} in controllable novel view synthesis. DC-Zero-1-to-3 effectively generates more controllable images from novel viewpoints by utilizing conditions from multi-view images.}
   \label{fig:zero123_nvs}
\end{figure*}

\input{table/zero123_nvs}

%% file: table/zero123_nvs.tex
\begin{table}
    \begin{minipage}{0.48\textwidth}
    \subcaption{Elevation Degree - 0}\label{subtab:ed0}
    \setlength{\tabcolsep}{12pt} 
    \renewcommand{\arraystretch}{1.08}
    \begin{tabular}{c|ccc}
        Methods & PSNR $\uparrow$ & SSIM $\uparrow$ & LPIPS $\downarrow$ \\
        \toprule
        Zero-1-to-3~\cite{liu2023zero1to3} & 20.82 & 0.840 & 0.139 \\
        Zero-1-to-3+Ours & \textbf{25.25} & \textbf{0.888} & \textbf{0.088} \\
    \end{tabular}
    \end{minipage} \\
    \begin{minipage}{0.48\textwidth}
    \subcaption{Elevation Degree - 15}\label{subtab:ed15}
    \setlength{\tabcolsep}{12pt} 
    \renewcommand{\arraystretch}{1.08}
    \begin{tabular}{c|ccc}
        Methods & PSNR $\uparrow$ & SSIM $\uparrow$ & LPIPS $\downarrow$ \\
        \toprule
        Zero-1-to-3 & 21.38 & 0.837 & 0.131 \\
        Zero-1-to-3+Ours & \textbf{25.85} & \textbf{0.891} & \textbf{0.083} \\
    \end{tabular}
    \end{minipage} \\
    \begin{minipage}{0.48\textwidth}
    \subcaption{Elevation Degree - 30}\label{subtab:ed30}
    \setlength{\tabcolsep}{12pt} 
    \renewcommand{\arraystretch}{1.08}
    \begin{tabular}{c|ccc}
        Methods & PSNR $\uparrow$ & SSIM $\uparrow$ & LPIPS $\downarrow$ \\
        \toprule
        Zero-1-to-3 & 21.66 & 0.837 & 0.128 \\
        Zero-1-to-3+Ours & \textbf{25.63} & \textbf{0.885} & \textbf{0.086} \\
    \end{tabular}
    \end{minipage} \\
    \caption{Quantitative analysis of novel view synthesis using the GSO dataset is presented, employing four distinct angles as inputs. The closest image to the desired viewpoint serves as the input for Zero-1-to-3 and the primary view for DC-Zero-1-to-3, with the remaining three images acting as auxiliary views for DC-Zero-1-to-3. Additionally, we compute results for input elevation angles set at 0, 15, and 30 degrees, respectively.}
    \label{tab:zero123_nvs}
\end{table}

%% file: tex/4_experiments.tex
\begin{figure*}
    \centering
    \includegraphics[width=1.0\linewidth]{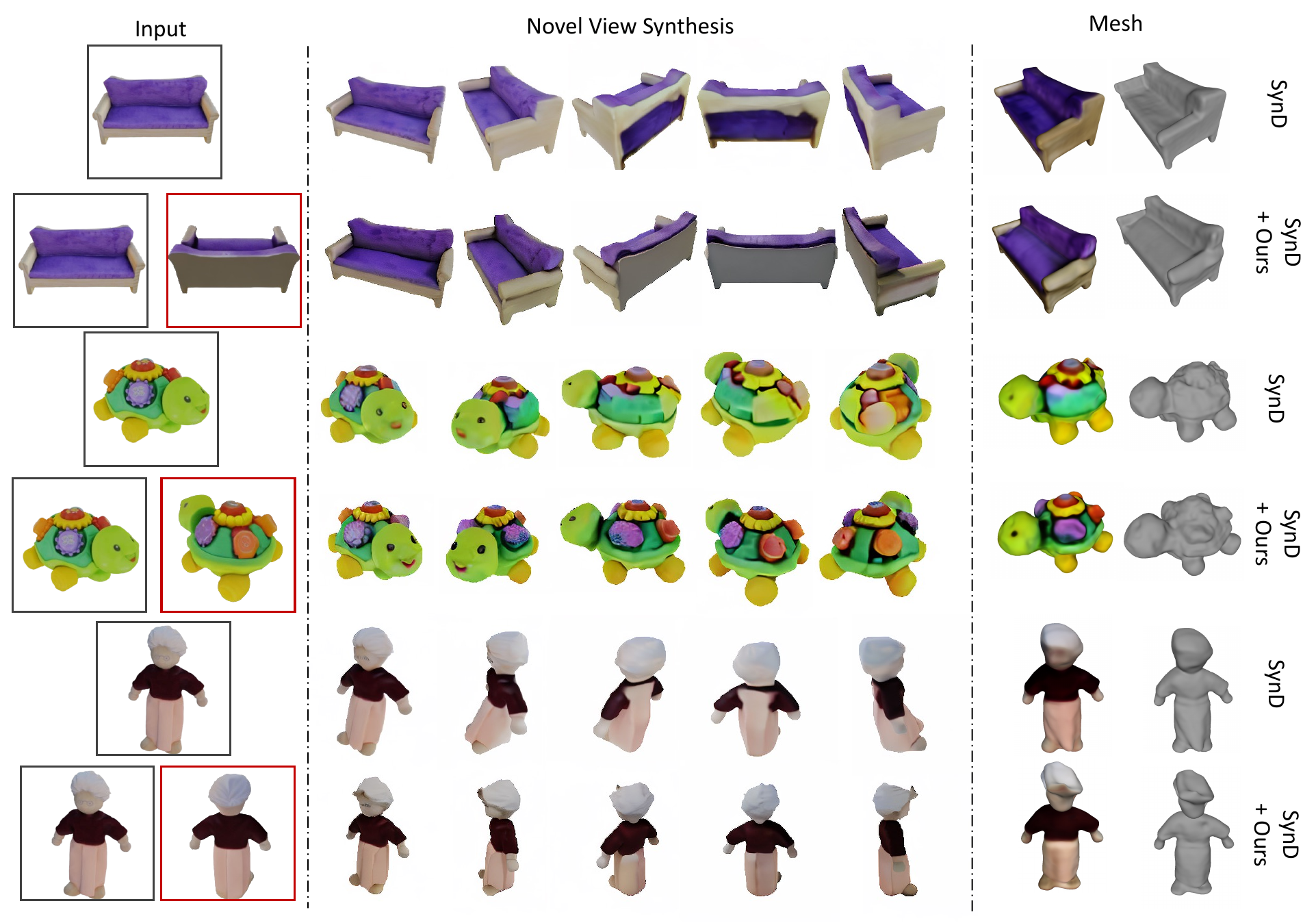}
    \caption{Qualitative comparison with SyncDreamer (SyncD)~\cite{liu2023syncdreamer} in controllable novel view synthesis and 3D reconstruction. The image in $\square$ is the main input, and the other image in \textcolor{red}{$\square$} is the conditional input generated from Zero-1-to-3~\cite{liu2023zero1to3}. With more information in multi-view images, DC-SyncDreamer is able to generate more accurate back textures and more controllable 3D shapes.}
    \label{fig:sync_nvs}
\end{figure*}

\begin{figure*}
    \centering
    \includegraphics[width=1.0\linewidth]{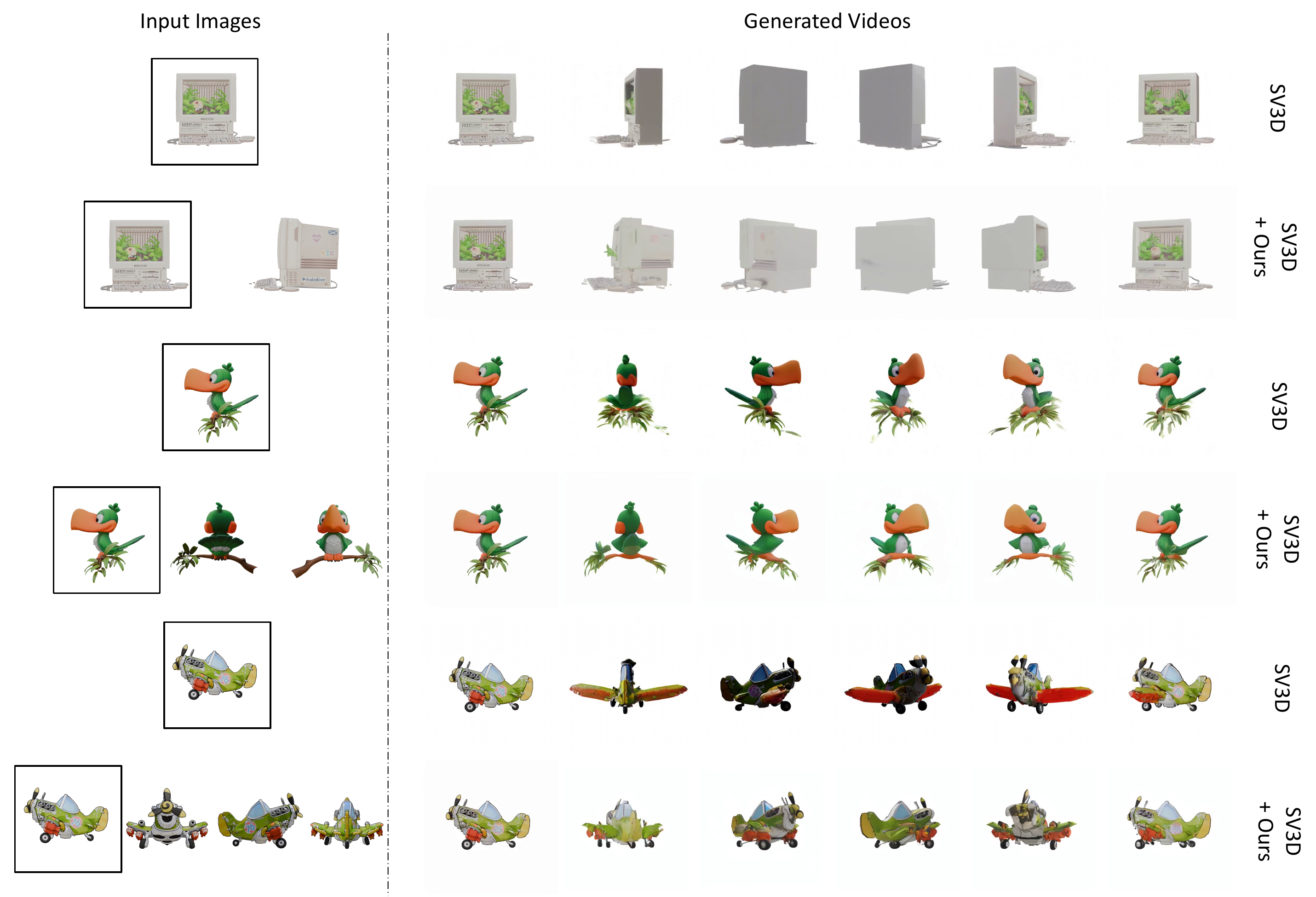}
    \caption{Qualitative comparison with SV3D in controllable novel view synthesis.}
    \label{fig:sv3d_nvs}
\end{figure*}

\input{table/syncdreamer_nvs}

\input{table/sv3d_nvs}

\begin{figure}
    \centering
    \includegraphics[width=1.0\linewidth]{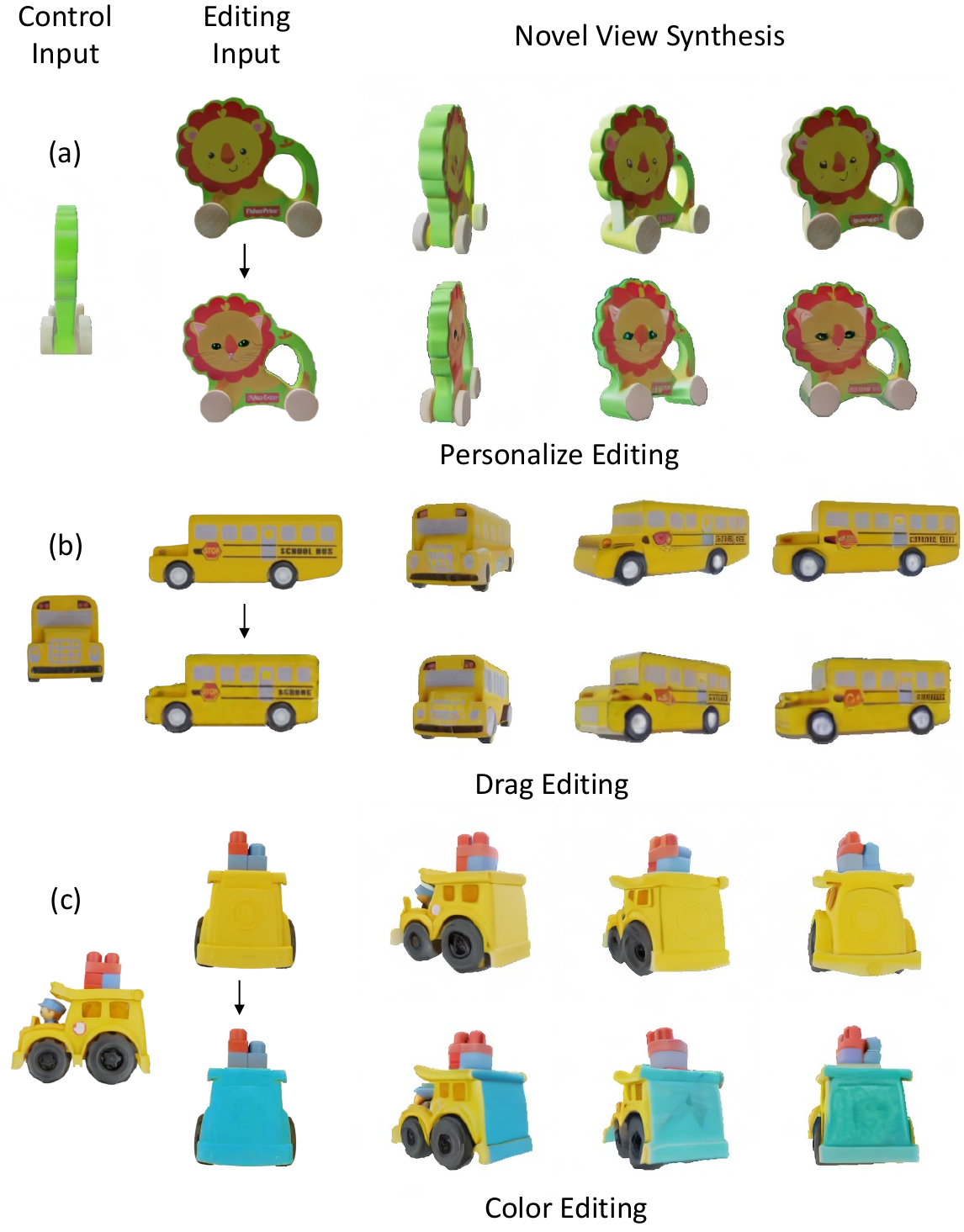}
    \caption{\textbf{Controllable Editing.} We present personalize editing with InstructPix2Pix~\cite{brooks2023instructpix2pix} in (a), drag editing with DragGAN~\cite{pan2023drag}, DragDiffusion~\cite{shi2023dragdiffusion} in (b), and color editing in (c).}
    \label{fig:editing}
\end{figure}

\begin{figure*}
    \centering
    \includegraphics[width=1.0\linewidth]{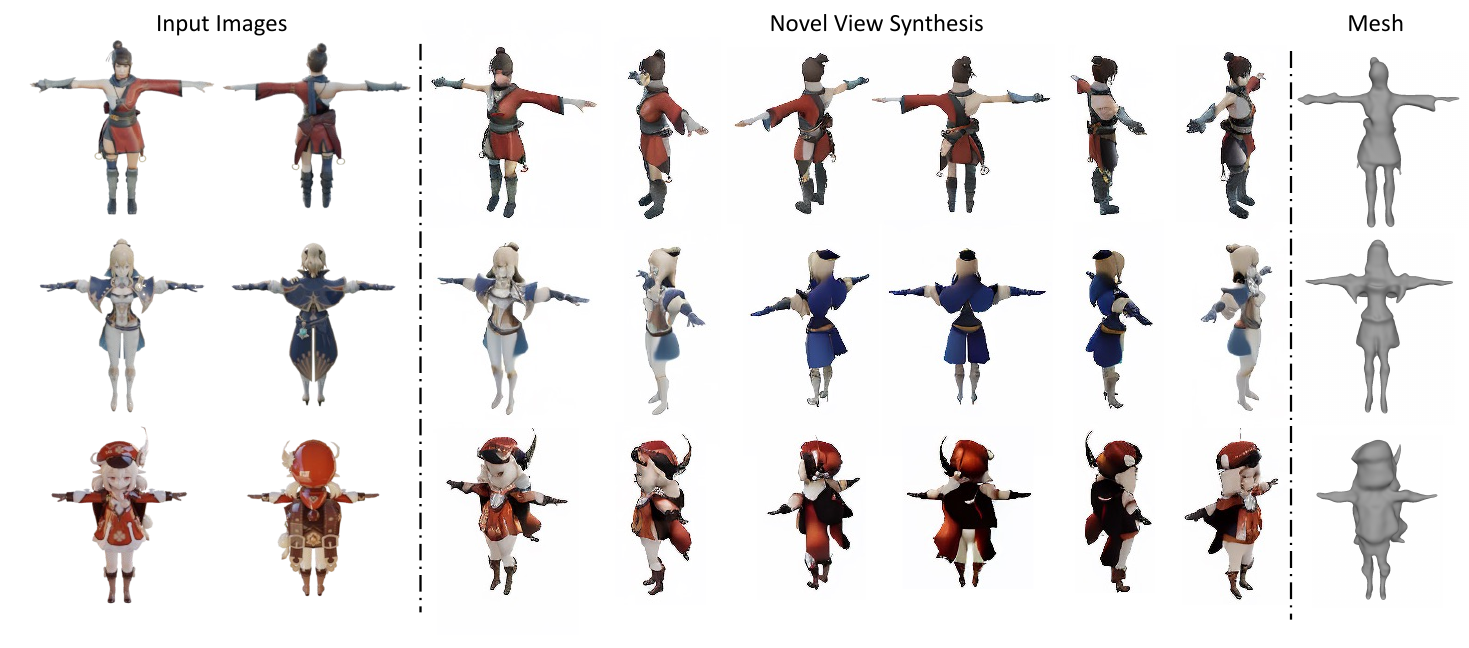}
    \caption{\textbf{3D Character Modeling using DC-SyncDreamer.} DC-SyncDreamer is able to reconstruct arbitrary objects with rarely sparse inputs. We present the results of 3D character modeling from multi-view 2D paintings.}
    \label{fig:control_recon}
\end{figure*}

\begin{figure*}
    \centering
    \includegraphics[width=1.0\linewidth]{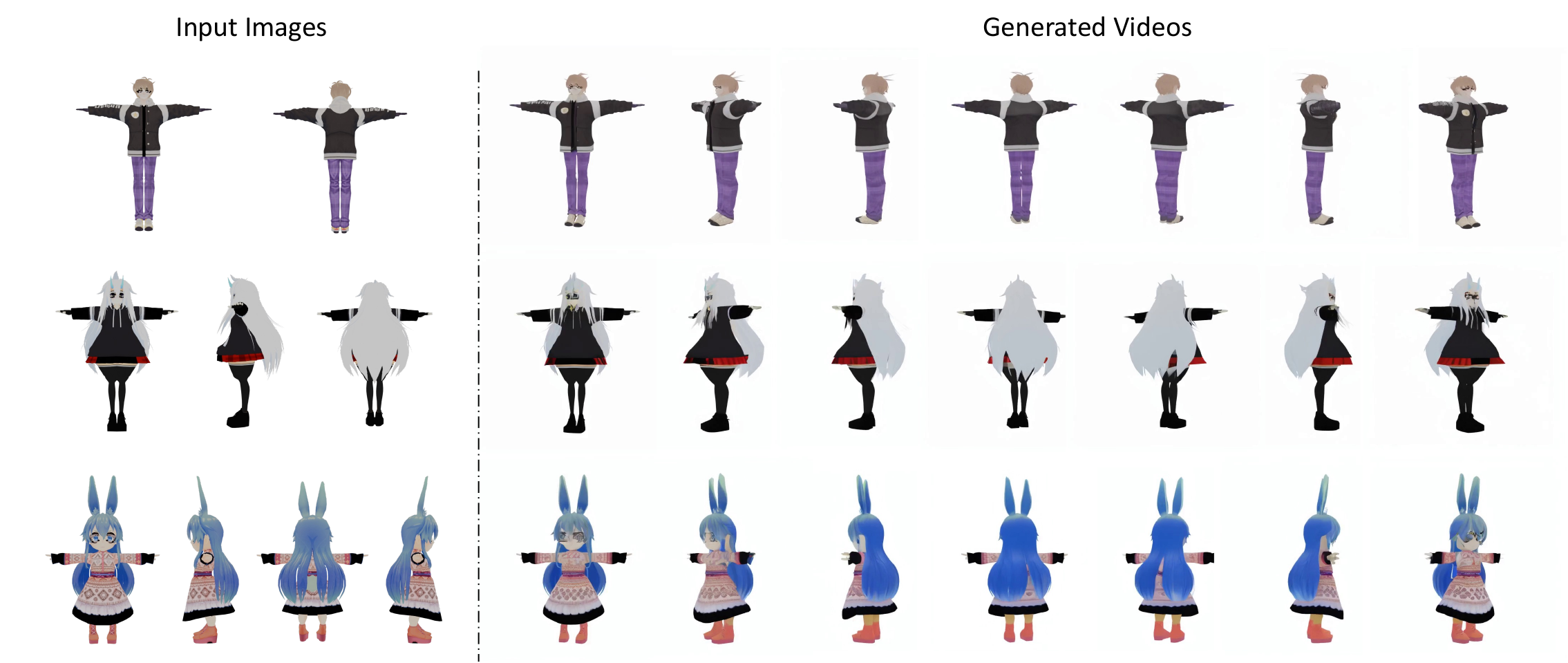}
    \caption{\textbf{3D Character Modeling using DC-SV3D.} DC-SV3D enables the reconstruction of 3D characters from arbitrary input views, providing enhanced texture control and higher resolution.}
    \label{fig:sv3d_character}
\end{figure*}

\begin{figure*}
  \centering
  \includegraphics[width=\linewidth]{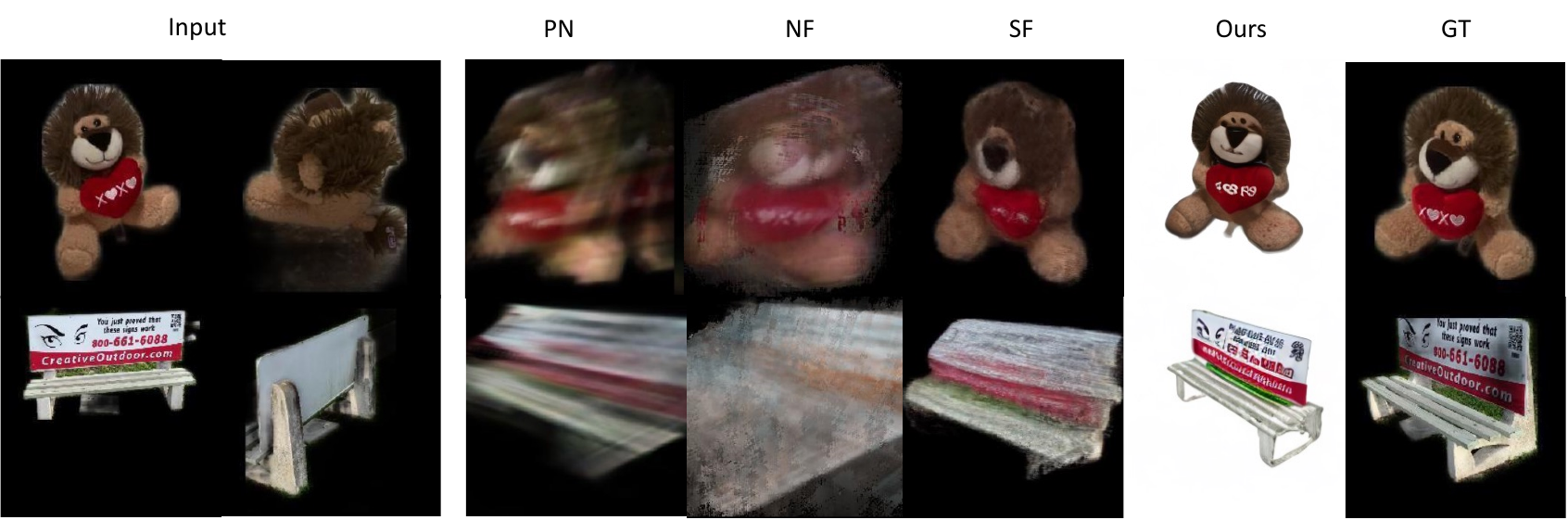}
   \caption{Qualitative comparison with PixelNeRF (PN), NerFormer (NF), SF (SparseFusion) in novel view synthesis.}
   \label{fig:co3d}
\end{figure*}

\begin{figure}
    \centering
    \includegraphics[width=1.0\linewidth]{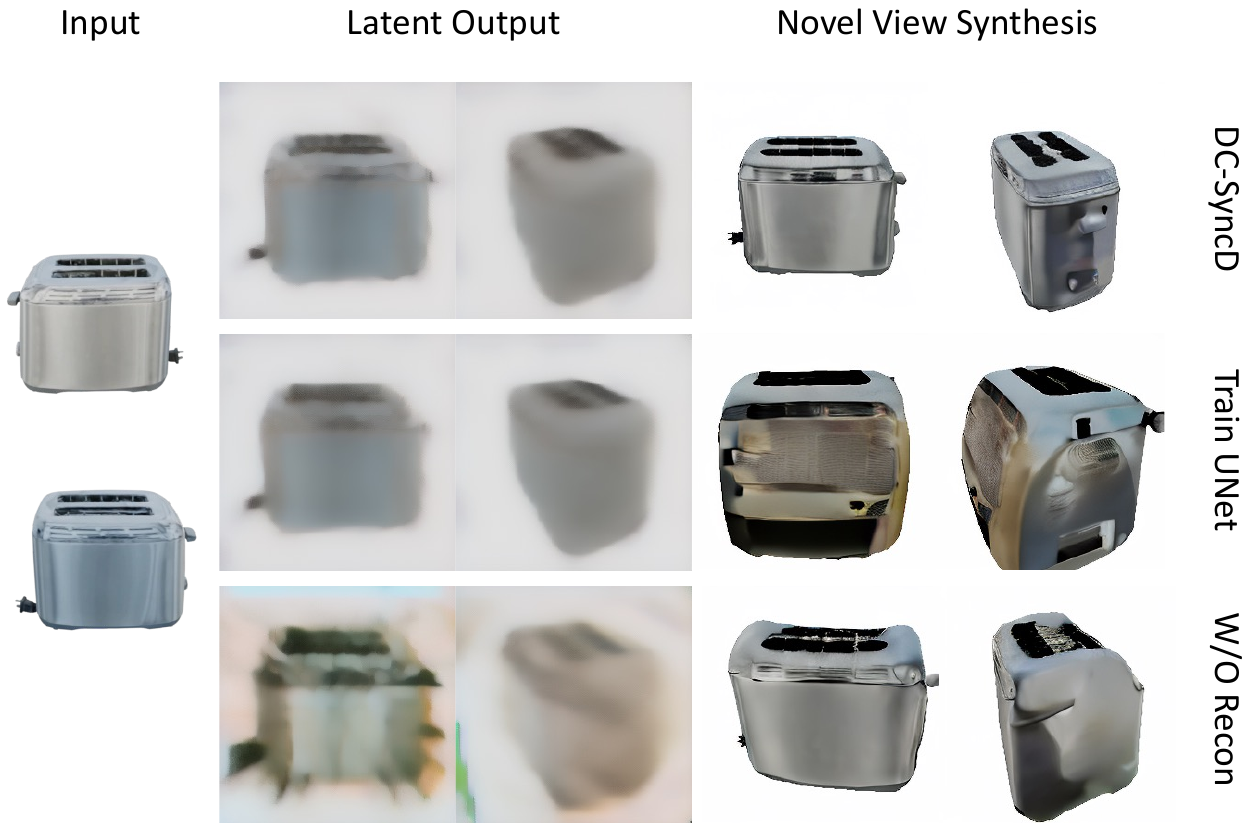}
    \caption{Ablation studies to verify the designs of our approach. ``DC-SyncD'' means our full model incorporating with SyncDreamer~\cite{liu2023syncdreamer} pipeline. ``Train UNet'' indicates finetuning the UNet with our modules without freezing it. ``W/O Recon'' means removing the reconstruction MSE loss in the second step of training. The Latent Output is derived by rendering and pooling features in the tri-planes, as shown in \figref{fig:pipeline}. 
    }
    \label{fig:ablation}
\end{figure}
\input{table/ablation}

\begin{figure*}
    \centering
    \includegraphics[width=1.0\linewidth]{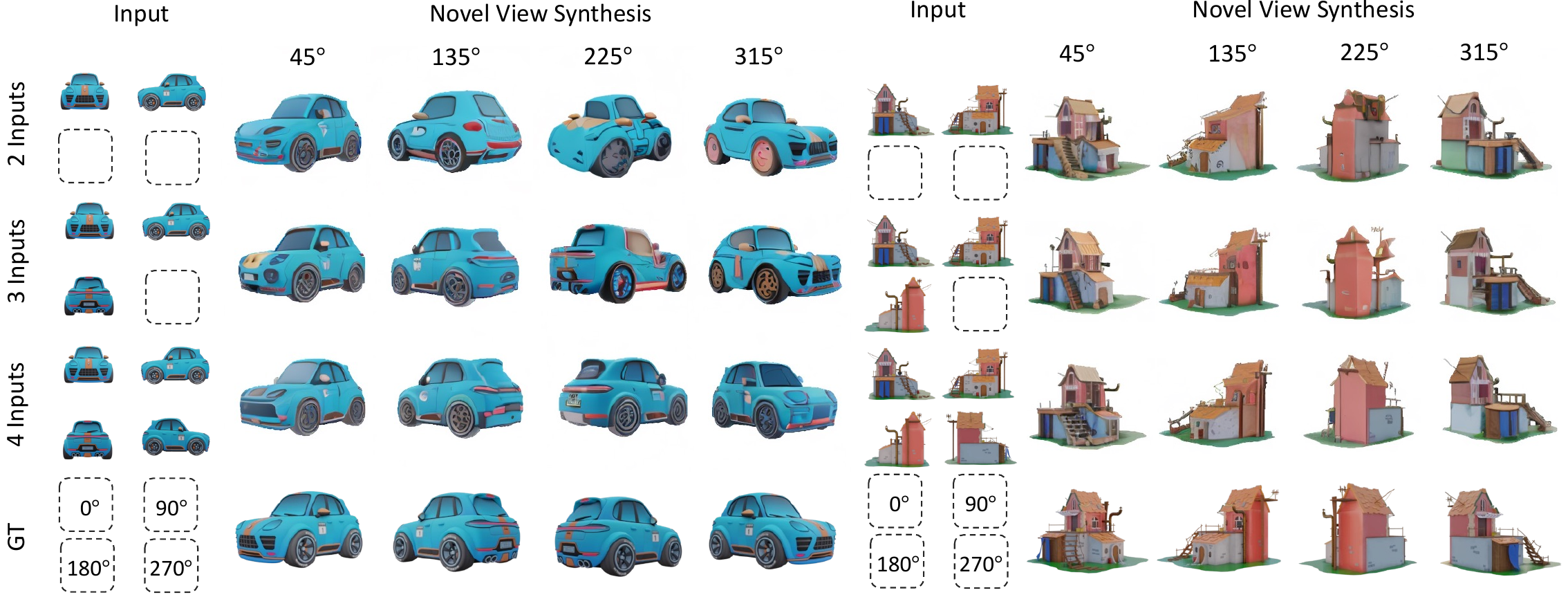}
    \caption{Ablation study to demonstrate the scalability of our model.
    Our model has the capacity to process arbitrary inputs, and its ability to control outcomes enhances correspondingly with the increasing information of input data.
    }
    \label{fig:ablation_flex}
\end{figure*}

\begin{figure*}
  \centering
  \includegraphics[width=\linewidth]{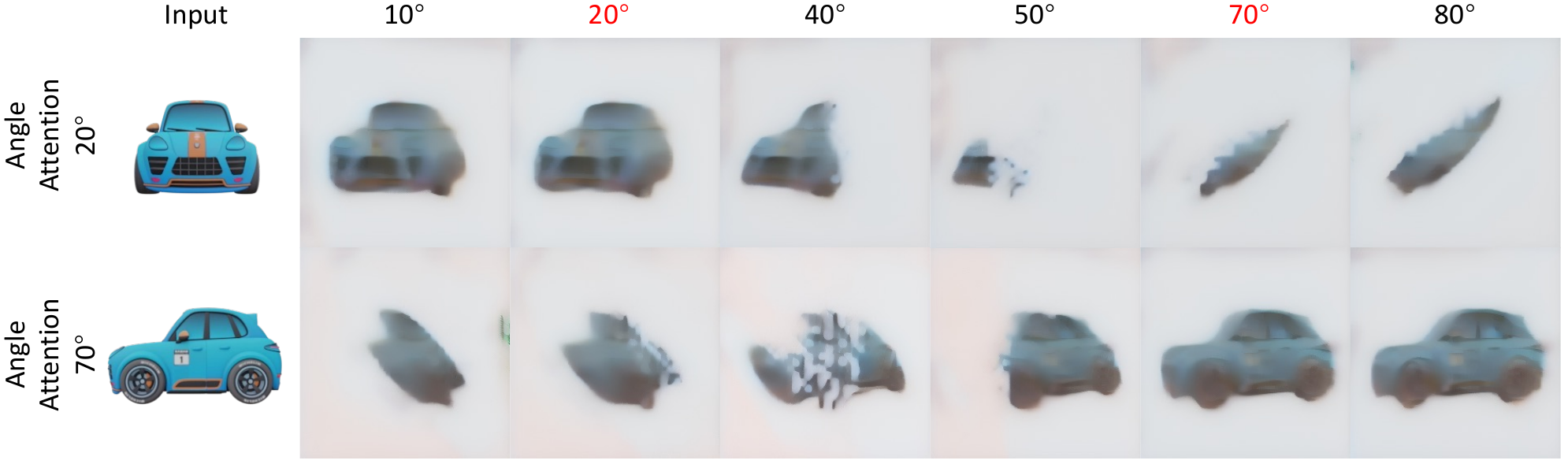}
   \caption{Latent space visualization with different angle attention. In the first row, a view difference of 20 degrees is specified, meaning only the result corresponding to 20 degrees is considered valid. Likewise, in the second row, the specified view difference is 70 degrees, so only the result at 70 degrees is valid. By using view-conditioning, the model can more effectively focus on enhancing the quality of the latent features for the target view.}
   \label{fig:angle_att}
   \vspace{-1em}
\end{figure*}

\input{table/ablation_scala}

\input{table/ablation_feature_fusion}

\section{Experiments}
\label{sec:exp}
We evaluate the effectiveness of DreamComposer++ on zero-shot novel view synthesis and 3D object reconstruction. The description of datasets and evaluation metrics are provided in \secref{sec:datasets}. Implementation details are presented in \secref{sec:imple_details}.
Our model is compatible with both image and video diffusion models. We integrate DreamComposer++ with image diffusion models Zero-1-to-3~\cite{liu2023zero1to3} and SyncDreamer~\cite{liu2023syncdreamer}, as described in \secref{sec:multi-view} and \secref{sec:single-view}, respectively. Additionally, we incorporate DreamComposer++ into the video diffusion model SV3D~\cite{voleti2024sv3d}, as introduced in \secref{sec:add_sv3d}. Comparisons with sparse view reconstruction methods are presented in \secref{sec:com_sparse}. Furthermore, we showcase the applications of DreamComposer++ in \secref{sec:application}, including controllable editing and 3D character modeling. The ablation study is introduced in \secref{sec:ablation}.

\subsection{Datasets and Evaluation Metrics}\label{sec:datasets}

\textbf{Training Dataset.}
We train DreamComposer++ (DC) on the large-scale Objaverse~\cite{deitke2022objaverse} dataset containing around 800k 3D objects. We randomly pick two elevation angles for every object and render $N$ images with the azimuth evenly distributed in $[0^\circ,360^\circ]$. We set $N$ to $36$ for DC-Zero-1-to-3, $16$ for DC-SyncDreamer and $N$ to $21$ for DC-SV3D. For training and inference, background is set to white.

\noindent\textbf{Evaluation Dataset.}
To evaluate the generalization of our model to out-of-distribution data, we extend our evaluation dataset from Objaverse to Google Scanned Objects (GSO)~\cite{downs2022google} and Common Objects In 3D (CO3Dv2)~\cite{reizenstein2021common}. The GSO dataset contains high-quality scans of everyday household items, whose evaluation setting is consistent with that for SyncDreamer~\cite{liu2023syncdreamer}, comprising 30 objects that include both commonplace items and various animal species. The CO3Dv2 dataset contains large-scale, in-the-wild 3D contents for evaluating novel view synthesis and 3D reconstruction tasks. We use CO3Dv2 to compare our approach with other sparse-view input methods.

\noindent\textbf{Evaluation Metrics.}
Following previous works~\cite{liu2023zero1to3, liu2023syncdreamer}, We utilize Peak Signal-to-Noise Ratio (PSNR), Structural Similarity Index (SSIM)~\cite{wang2004image}, and Learned Perceptual Image Patch Similarity (LPIPS)~\cite{zhang2018unreasonable} as our evaluation metrics.

\subsection{Implementation Details}\label{sec:imple_details}

\noindent\textbf{Camera Embedding.} Following Zero-1-to-3~\cite{liu2023zero1to3}, we utilize a spherical coordinate system to represent camera locations and their relative transformations. As shown in \figref{fig:cam_system}, during the training stage, camera locations of two images from disparate viewpoints are designated as $(\theta_1, \phi_1, r_1)$ and $(\theta_2, \phi_2, r_2)$, respectively. The \emph{relative} transformation between these camera positions is expressed as $(\theta_2 - \theta_1, \phi_2 - \phi_1, r_2 - r_1)$. In both the training and inference stages, four parameters delineating the relative camera viewpoint $[\Delta\theta, \sin(\Delta\phi), \cos(\Delta\phi), \Delta r]$ are inputted into the cross-attention layers of DreamComposer++'s Target-Aware 3D Lifting Module and Target-View Feature Injection Module to provide camera view information.

\noindent\textbf{Architecture and Hyperparameters.} We design Target-Aware 3D Lifting Module based on the U-Net architecture from Stable Diffusion~\cite{rombach2021highresolution}. This model's architecture is specifically configured with a model dimension of 192 and includes two residual blocks at each resolution level. A distinctive feature of our approach is the integration of a cross-attention module, which facilitates the processing of relative camera embeddings.

For our experiments with Zero-1-to-3~\cite{liu2023zero1to3} and SyncDreamer~\cite{liu2023syncdreamer}, we standardize the image dimensions at $256 \times 256$ pixels. Correspondingly, this establishes the latent space dimensionality at $32 \times 32$. Additionally, we configure the triplane dimensions at $32 \times 32 \times 3$, with the feature dimension of each triplane element being set to 32. And for experiments with SV3D~\cite{voleti2024sv3d}, we set image dimensions as $576 \times 576$ pixels, latent space dimensions as $72 \times 72$.

\noindent\textbf{Training Details.} We adopt a two-stage training strategy for DreamComposer++. In the first stage, we focus on the 3D feature lifting module and pre-train it for 80k steps ($\sim$ 3 days) with 8 80G A800 GPUs using a total batch size of 576. The pre-trained 3D lifting module can be applied in conjunction with different pre-trained diffusion models for subsequent training. In the second stage, we jointly optimize the 3D lifting and feature injection module. This stage takes 30k steps ($\sim$ 2 days) with 8 80G A800 GPUs using a total batch size of 384 for DC-Zero-1-to-3 and DC-SyncDreamer, and 50k steps with 4 80G A800 GPUs using a total batch size of 4 for DC-SV3D.

During the entire training process, we randomly pick a target image as ground truth and utilize a set of two or three images as inputs: two images captured from opposing angles and optional one image from a random angle. Benefited from the image triplet training scheme, our model can adapt to two or more inputs. This data sampling strategy not only improves the efficiency of the model's optimization but also preserves its scalability and adaptability to various input configurations.

\subsection{Plugged into Zero-1-to-3}\label{sec:multi-view}

In this section, we evaluate the performance of DreamComposer++ plugged into the Zero-1-to-3 with multi-view ground-truth inputs, as depicted in \figref{fig:input-num-vis} (a). 

\noindent\textbf{Evaluation Protocols.} 
When provided with an input image of an object, Zero-1-to-3~\cite{liu2023zero1to3} has the ability to generate new perspectives of the same object. 
We take the four orthogonal angles as input and set the image closest to the target perspective as the input for Zero-1-to-3 as well as the main view for DC-Zero-1-to-3. The remaining three images serve as the additional condition-views for DC-Zero-1-to-3. We calculate the results for input elevation angles of 0, 15, and 30 degrees respectively.

\noindent\textbf{Evaluation on NVS.} The comparison of quantitative results is shown in \tabref{tab:zero123_nvs}, and the comparison of qualitative results is shown in \figref{fig:zero123_nvs}. While Zero-1-to-3 possesses the ability to produce visually plausible images from novel views, the absence of multi-view inputs compromises the accuracy of these unseen viewpoints. Our DC-Zero-1-to-3, by conditioning on multi-view images, ensures the controlled generation of new viewpoints while maintaining the integrity of its diffusion model's generative capabilities. DC-Zero-1-to-3 significantly surpasses other methods in terms of the quality and consistency of generated images across various angles.

\subsection{Plugged into SyncDreamer}\label{sec:single-view}
In this section, we evaluate the performance of DreamComposer++ plugged into the SyncDreamer~\cite{liu2023syncdreamer} with single-view ground-truth inputs, as depicted in \figref{fig:input-num-vis} (b).

\noindent\textbf{Evaluation Protocols.} We compare our method with SyncDreamer~\cite{liu2023syncdreamer}, Zero-1-to-3~\cite{liu2023zero1to3}, and RealFusion~\cite{melas2023realfusion}. Given an input image of an object, Zero-1-to-3 can synthesize novel views of the object, and SyncDreamer is able to generate consistent novel views from 16 fixed views. 
RealFusion~\cite{melas2023realfusion} is a single-view reconstruction method based on Stable Diffusion~\cite{rombach2021highresolution} and SDS~\cite{poole2022dreamfusion}. The inverse perspective of the input, generated using Zero-1-to-3~\cite{liu2023zero1to3}, serves as an additional condition-view for DC-SyncDreamer. We adhere to the identical input configurations as established in SyncDreamer. The mesh is directly reconstructed from multi-view images by NeuS~\cite{wang2023neus}. 

\noindent\textbf{Evaluation on NVS and 3D Reconstruction.} The comparison of quantitative results is shown in \tabref{tab:sync_nvs}, and the comparison of qualitative results is shown in \figref{fig:sync_nvs}. While SyncDreamer is able to generate consistent novel views, the shape of the object and the texture on the back may still appear unreasonable. DC-SyncDreamer not only maintains multi-view consistency in colors and geometry but also enhances the control over the shape and texture of the newly generated perspectives.

\subsection{Plugged into SV3D}\label{sec:add_sv3d}
In this section, we evaluate the performance of DreamComposer++ plugged into the video diffusion model SV3D~\cite{voleti2024sv3d}.

\noindent\textbf{Evaluation Protocols.} Given a single-perspective image of an object, SV3D can generate a consistent 21-frame video representing novel views of the object. While the generated video can be used to reconstruct the 3D object, it is not controllable due to the limited information from the single view. To assess the effectiveness of our model in enhancing SV3D, we use 2, 3, 4, 5, or 6 views as input to generate novel view videos.

\noindent\textbf{Evaluation on NVS.} The quantitative comparison is presented in \tabref{tab:sv3d_nvs}. With more input views, the videos generated by DC-SV3D are significantly more controllable than those generated by SV3D, particularly as reflected in the LPIPS metric. Additionally, as the number of input images increases, the generated videos become increasingly controllable. The qualitative comparison is shown in \figref{fig:sv3d_nvs}, which includes SV3D's generated videos using 2, 3, and 4 input views. With scalable DreamComposer++, DC-SV3D is able to generate controllable, high-quality videos representing novel views of 3D content.

\subsection{Comparison with Sparse View Reconstruction Methods}\label{sec:com_sparse}
To quantitatively compare with novel view synthesis methods from sparse-view inputs, we choose ViewFormer~\cite{kulhanek2022viewformer} as a competitor, as it achieves significant results on the CO3D dataset~\cite{reizenstein2021common}.
ViewFormer is designed for novel view synthesis using sparse-view inputs, employing transformers to process multiple context views and a query pose. This approach allows for the synthesis of novel images within an advanced neural network architecture.  For our evaluation, we utilize the ViewFormer model that has been comprehensively trained on the CO3D dataset~\cite{reizenstein2021common}, ensuring a fair comparison with its contemporary counterparts. The evaluation dataset setting is same as the one in \secref{sec:multi-view}. 
The quantitative results are shown in \tabref{tab:zero123_nvs_supp}. While ViewFormer shows proficiency in handling the CO3D dataset, it exhibits limitations in processing out-of-distribution data.
\input{table/zero123_nvs_supp}

Our method is capable of zero-shot learning and also demonstrates superior performance compared to other few-shot reconstruction methods when testing on their datasets. Since the contents in CO3DV2 dataset is not at the center of images, it does not align to the setting of Zero-1-to-3 and SyncDreamer, making it unsuitable for quantitative evaluation. However, we provide a qualitative comparison with PixelNeRF (PN)~\cite{yu2021pixelnerf}, NerFormer (NF)~\cite{reizenstein2021common} and SF (SparseFusion)~\cite{zhou2023sparsefusion} in \figref{fig:co3d}.
DC-Zero-1-to-3 significantly outperforms other novel view synthesis methods with sparse-view inputs in both qualitative and quantitative analyses.

\subsection{Applications}\label{sec:application}
We explore the various applications of DreamComposer++, including controllable 3D object editing with DC-Zero-1-to-3 and 3D character modeling with DC-SyncDreamer.

\noindent\textbf{Controllable 3D object Editing.} DreamComposer++ is able to perform controllable editing by modifying or designing images from certain perspectives, as shown in \figref{fig:editing}. We designate an image from a specific viewpoint as the ``control input'', which remains unaltered. Concurrently, we manipulate an ``editing input'', which represents an image from an alternate viewpoint. We utilize InstructPix2Pix~\cite{brooks2023instructpix2pix}, DragGAN~\cite{pan2023drag} and DragDiffusion~\cite{shi2023dragdiffusion} to manipulate the image, thereby achieving our desired style, corresponding to (a), (b) in \figref{fig:editing} respectively. And we modify the color of the editing input ourselves in (c). Subsequently, we employ the modified images in conjunction with the control input to synthesize novel views.

\noindent\textbf{3D Character Modeling.}
DC-SyncDreamer allows 3D character modeling from just a few 2D paintings, as illustrated in \figref{fig:control_recon}. By leveraging DC-SV3D, we can model 3D characters from an arbitrary number of input views, enabling finer texture control and higher resolution, as shown in \figref{fig:sv3d_character}. This can significantly improve the efficiency of existing 3D pipelines, and is expected to be connected with ControlNet~\cite{zhang2023adding} for text-to-3D character creation~\cite{huang2023dreamwaltz}.

\subsection{Ablation Analysis}\label{sec:ablation}
We conduct ablation studies on DC-SyncDreamer. For the ablation study, quantitative results are included in \tabref{tab:ablation} and qualitative samples are included in \figref{fig:ablation}.

\noindent\textbf{Necessity of reconstruction loss.} First, we remove the reconstruction MSE loss in the second step of training as discussed in \secref{sec:multi_view_training}. As shown in \figref{fig:ablation} and \tabref{tab:ablation}, without reconstruction MSE loss, the multi-view 3D lifting module is unable to produce effective latent outputs, resulting in the inability to synthesize satisfactory novel views.

\noindent\textbf{Finetuning v.s. freezing the diffusion U-Net.} In our design, the pretrained diffusion U-Net is frozen during training DreamController. We attempt to finetune the diffusion U-Net with DreamComposer++'s modules in the second stage of training. As shown in \figref{fig:ablation} and \tabref{tab:ablation}, the model performance decreases when we finetune the U-Net together with our modules. 

\noindent\textbf{Necessity of view-conditioning for 3D lifting.} Quantitatively, we evaluate the impact of removing the view-conditioning cross-attention from the 2D-to-3D lifting module. As shown in \tabref{tab:ablation}, this removal results in degraded performance.

Qualitatively, we visualize the tri-plane features from the target view under different angular differences. As shown in \figref{fig:angle_att}, the projection from the target view exhibits the highest quality. The experimental setup, described in the first column, involves two inputs, with the primary view shown at the top. The view difference is calculated relative to this primary view. In the first row, the specified view difference is 20 degrees, so only the result corresponding to 20 degrees is considered valid. Similarly, in the second row, a view difference of 70 degrees is specified, making only the result at the corresponding 70 degrees valid. With view-conditioning, the model can more effectively focus on improving the quality of the latent features for the target view.

\noindent\textbf{Scalability for arbitrary numbers of input views.} We validate our model's flexibility and scalability in managing arbitrary numbers of inputs. As shown in \figref{fig:ablation_flex}, our model can handle arbitrary numbers of input views, and its controllability is strengthened proportionally with the increasing number of input views. The quantitative results of DC-Zero-1-to-3 with varying numbers of input views are shown in \tabref{tab:ablation_scala}.

\begin{figure}
    \centering
    \includegraphics[width=1.0\linewidth]{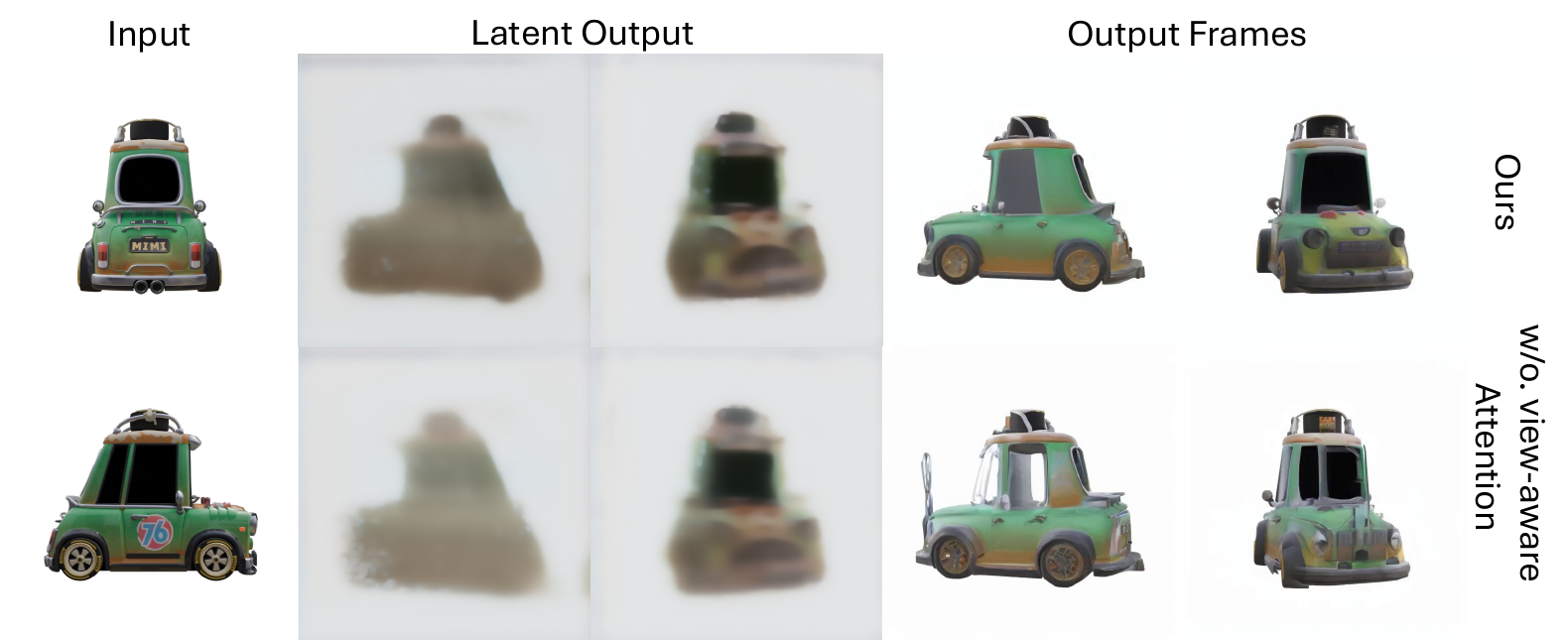}
    \caption{Comparison of video generation with and without View-aware Attention.}
    \label{fig:ablation_attn}
\end{figure}

\textbf{Different design of feature controlling module.} We introduce View-aware Attention and Shifted Window Cross-attention into the feature controlling module to enhance multi-view consistency, control, and robustness. A quantitative comparison in \tabref{tab:abl_fusion} shows that our full model significantly outperforms the variants without these components across all elevation angles, with improvements in PSNR, SSIM, and LPIPS metrics. Specifically, the absence of View-aware Attention leads to a substantial drop in SSIM and a notable increase in LPIPS, indicating weakened structural fidelity and perceptual quality. Furthermore, replacing our attention-based design with a simple convolutional fusion also results in consistently lower scores, suggesting that convolution alone is insufficient for maintaining inter-view coherence. In addition, the qualitative comparison in \figref{fig:ablation_attn} highlights the visual benefits of our proposed modules. Models without View-aware Attention often exhibit inconsistent shading, geometric distortion, or view-dependent artifacts when rendered from different angles. In contrast, our approach produces more stable geometry and coherent textures across views, better reflecting realistic 3D structures. These results validate the importance of incorporating cross-view attention mechanisms into the feature fusion pipeline for effective multi-view generation.

\begin{figure}
    \centering
    \includegraphics[width=1.0\linewidth]{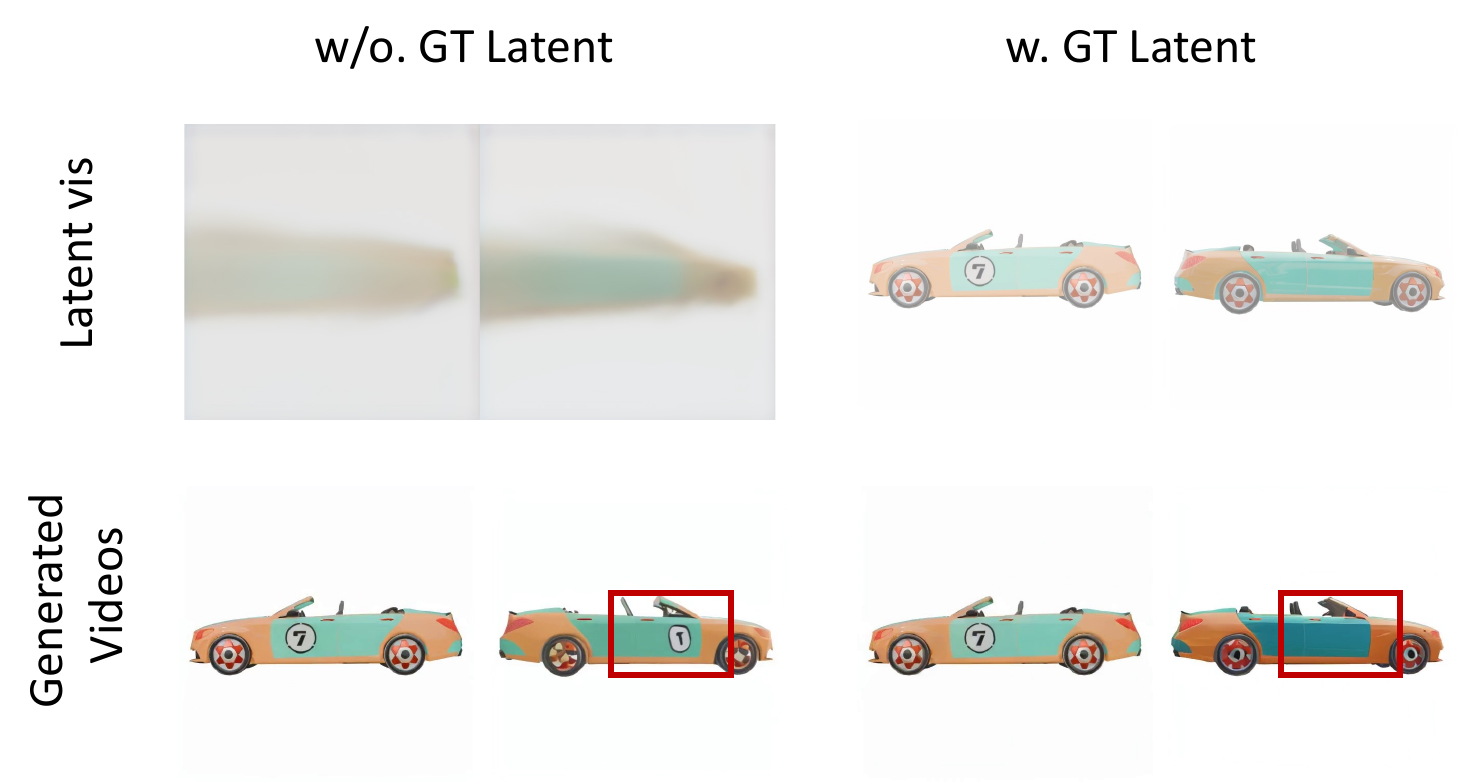}
    \caption{Comparison of video generation w/o and w/ Ground Truth latents. Using GT latents (right) leads to sharper and more consistent results compared to Lift3D outputs (left).}
    \label{fig:ablation_lift}
\end{figure}

\noindent\textbf{Bottle Neck Analysis.} We utilize the Ground Truth image latents as input in place of the 3D features rendered by Lift3D to guide the hint module. As illustrated in the comparison Figure \ref{fig:ablation_lift}, the visual quality and control over the generated video significantly improve when using Ground Truth latents, whereas the results obtained with Lift3D outputs show weaker control and less coherence. This discrepancy is largely due to the current limitations of Lift3D, which struggles to generate high-quality latent features, thus impacting the downstream video generation process.

When Lift3D-rendered features are used, the model's ability to maintain consistency and produce accurate video frames diminishes, leading to suboptimal control over the generated content. On the contrary, when using Ground Truth latents, the method demonstrates strong control and temporal consistency, resulting in high-quality video outputs. These findings suggest that the primary bottleneck in improving the entire framework lies in enhancing the Lift3D module. By boosting Lift3D’s capacity to generate higher-quality 3D representations, the overall performance of the video generation process would significantly improve, allowing for better control and more reliable outputs across different conditions.

%% file: table/syncdreamer_nvs.tex
\begin{table}
    \centering
    \setlength{\tabcolsep}{12pt} 
    \renewcommand{\arraystretch}{1.08}
    \begin{tabular}{c|ccc}
       Method  & PSNR$\uparrow$ & SSIM$\uparrow$ & LPIPS$\downarrow$ \\
       \toprule
       Realfusion~\cite{melas2023realfusion}    
       & 15.26 & 0.722 & 0.283   \\
       Zero-1-to-3~\cite{liu2023zero1to3}    
       & 18.93 & 0.779 & 0.166   \\
       SyncDreamer~\cite{liu2023syncdreamer}    
       & 20.05 & 0.798 & 0.146   \\
       SyncDreamer+Ours 
       & \textbf{20.52} & \textbf{0.828} & \textbf{0.141}   \\
    \end{tabular}
    \caption{Quantitative comparisons of novel view synthesis on GSO dataset. We employ images generated from diffusion models as our additional condition-view.}
    \label{tab:sync_nvs}
\end{table}

%% file: table/sv3d_nvs.tex
\begin{table*}
    \centering
    \begin{tabularx}{0.95\textwidth}{c|XXX|XXX|XXX}
       & \multicolumn{3}{c|}{Elevation Degree - 0} & \multicolumn{3}{c|}{Elevation Degree - 15} & \multicolumn{3}{c}{Elevation Degree - 30} \\
        & PSNR$\uparrow$ & SSIM$\uparrow$ & LPIPS$\downarrow$ & PSNR$\uparrow$ & SSIM$\uparrow$ & LPIPS$\downarrow$ & PSNR$\uparrow$ & SSIM$\uparrow$ & LPIPS$\downarrow$ \\
       \toprule
        SV3D & 16.98 & 0.8419 & 0.3156  & 16.67 &0.7677& 0.3005&16.34&0.7539&0.3127\\

       
        SV3D+Ours (2 views) & 19.83 & 0.8347 & 0.1441 & 19.44 & 0.8177& 0.1494& 19.78& 0.8138& 0.1517 \\

        SV3D+Ours (3 views) & 21.32 & 0.8548 & 0.1198 & 21.53 & 0.8470 & 0.1225 & 21.18 & 0.8325 & 0.1343 \\
       
        SV3D+Ours (4 views) & 21.73 & 0.8607 & 0.1131 & 21.94 & 0.8535 & 0.1163 & 21.62 & 0.8389 & 0.1279 \\

        SV3D+Ours (5 views) & 21.99 & 0.8650 & 0.1093 & 22.03 & 0.8552 & 0.1151 & 21.72 & 0.8409 & 0.1262\\
       
        SV3D+Ours (6 views) & \textbf{22.21} & \textbf{0.8686} & \textbf{0.1067} & \textbf{22.21} & \textbf{0.8586} & \textbf{0.1128} & \textbf{21.80} & \textbf{0.8430} & \textbf{0.1244}  \\


    \end{tabularx}
    \caption{Quantitative comparisons with SV3D on the GSO dataset with different number of inputs. As the number of input images increases, the generation of new perspectives becomes more controllable.}
    \label{tab:sv3d_nvs}
\end{table*}

%% file: table/ablation.tex
\begin{table}
    \centering
    \setlength{\tabcolsep}{12pt}
    \renewcommand{\arraystretch}{1.08}
    \setlength{\tabcolsep}{11pt}
    \begin{tabular}{c|ccc}
       Method  & PSNR$\uparrow$ & SSIM$\uparrow$ & LPIPS$\downarrow$ \\
       \toprule
       trainable UNet
       & 15.96 & 0.762 & 0.209   \\
       w/o reconstruction loss
       & 16.18 & 0.766 & 0.206   \\
       w/o view conditioning
       & 19.04 & 0.805 & 0.166   \\
       full model 
       & \textbf{20.52} & \textbf{0.828} & \textbf{0.141}   \\
    \end{tabular}
    \caption{Ablation study on GSO dataset. Eliminating the reconstruction loss and training the UNet are both factors that negatively impact the final outcome. With view conditioning in the 3D lifting module, our model not only ensures more stable training but also yields the most optimal results.}
    \label{tab:ablation}
\end{table}

%% file: table/ablation_scala.tex
\begin{table*}
    \centering
    \begin{tabularx}{0.95\textwidth}{c|XXX|XXX|XXX}
       & \multicolumn{3}{c|}{Elevation Degree - 0} & \multicolumn{3}{c|}{Elevation Degree - 15} & \multicolumn{3}{c}{Elevation Degree - 30} \\
        & PSNR$\uparrow$ & SSIM$\uparrow$ & LPIPS$\downarrow$ & PSNR$\uparrow$ & SSIM$\uparrow$ & LPIPS$\downarrow$ & PSNR$\uparrow$ & SSIM$\uparrow$ & LPIPS$\downarrow$ \\
       \toprule
        2 views & 20.38 & 0.826 & 0.159 & 22.33 & 0.847 & 0.125 & 22.42 & 0.845 & 0.124 \\

        3 views & 23.68 & 0.869 & 0.108 & 24.56 & 0.875 & 0.098 & 24.27 & 0.867 & 0.102 \\
       
        4 views & 25.25 & 0.888 & 0.088 & 25.85 & 0.891 & 0.083 & 25.63 & 0.885 & 0.086 \\

        5 views & 26.10 & 0.897 & 0.081 & 26.62 & 0.899 & 0.078 & 26.52 & 0.895 & 0.079 \\
       
        6 views & \textbf{26.99} & \textbf{0.907} & \textbf{0.074} & \textbf{27.39} & \textbf{0.907} & \textbf{0.072} & \textbf{27.26} & \textbf{0.903} & \textbf{0.073}
    \end{tabularx}
    \caption{Quantitative comparisons on the GSO dataset with different number of inputs. As the number of input images increases, the generation of new perspectives becomes more controllable.}
    \label{tab:ablation_scala}
\end{table*}

%% file: table/ablation_feature_fusion.tex
        



       

\begin{table*}
    \centering
    \begin{tabularx}{0.95\textwidth}{c|XXX|XXX|XXX}
       & \multicolumn{3}{c|}{Elevation Degree - 0} & \multicolumn{3}{c|}{Elevation Degree - 15} & \multicolumn{3}{c}{Elevation Degree - 30} \\
        & PSNR$\uparrow$ & SSIM$\uparrow$ & LPIPS$\downarrow$ & PSNR$\uparrow$ & SSIM$\uparrow$ & LPIPS$\downarrow$ & PSNR$\uparrow$ & SSIM$\uparrow$ & LPIPS$\downarrow$ \\
       \toprule

      Ours & \textbf{19.83} & \textbf{0.8347} & \textbf{0.1441} & \textbf{19.44} & \textbf{0.8177}& \textbf{0.1494}& \textbf{19.78}& \textbf{0.8138}& \textbf{0.1517} \\
        
        W/o View-aware Attn. & 17.61 & 0.8193 & 0.2968 & 17.69 & 0.7940& 0.2927 & 17.57 & 0.7567& 0.3066\\

        Conv Fusion & 17.98 & 0.8321 & 0.1723 & 17.88 & 0.7942& 0.1764 & 17.00 & 0.7734& 0.2044\\



       
    \end{tabularx}
    \caption{Ablation analysis of the design of View-aware Attention and Shifted Window Cross-attention for better incorporation with video diffusion models.}
    \label{tab:abl_fusion}
\end{table*}

%% file: table/zero123_nvs_supp.tex
\begin{table}
    \begin{minipage}{0.48\textwidth}
    \subcaption{Elevation Degree - 0}\label{subtab:ed0_supp}
    \setlength{\tabcolsep}{12pt} 
    \renewcommand{\arraystretch}{1.08}
    \begin{tabular}{c|ccc}
        Methods & PSNR $\uparrow$ & SSIM $\uparrow$ & LPIPS $\downarrow$ \\
        \toprule
        ViewFormer & 13.45 & 0.630 & 0.359 \\
        DC-Zero-1-to-3 (Ours) & \textbf{25.25} & \textbf{0.888} & \textbf{0.088} \\
    \end{tabular}
    \end{minipage} \\
    \begin{minipage}{0.48\textwidth}
    \subcaption{Elevation Degree - 15}\label{subtab:ed15_supp}
    \setlength{\tabcolsep}{12pt} 
    \renewcommand{\arraystretch}{1.08}
    \begin{tabular}{c|ccc}
        Methods & PSNR $\uparrow$ & SSIM $\uparrow$ & LPIPS $\downarrow$ \\
        \toprule
        ViewFormer & 13.00 & 0.618 & 0.371 \\
        DC-Zero-1-to-3 (Ours) & \textbf{25.85} & \textbf{0.891} & \textbf{0.083} \\
    \end{tabular}
    \end{minipage} \\
    \begin{minipage}{0.48\textwidth}
    \subcaption{Elevation Degree - 30}\label{subtab:ed30_supp}
    \setlength{\tabcolsep}{12pt} 
    \renewcommand{\arraystretch}{1.08}
    \begin{tabular}{c|ccc}
        Methods & PSNR $\uparrow$ & SSIM $\uparrow$ & LPIPS $\downarrow$ \\
        \toprule
        ViewFormer & 13.02 & 0.618 & 0.373 \\
        DC-Zero-1-to-3 (Ours) & \textbf{25.63} & \textbf{0.885} & \textbf{0.086} \\
    \end{tabular}
    \end{minipage} \\
    \caption{Quantitative comparisons of novel view synthesis on 
    GSO dataset using four orthogonal angles' images as inputs.}
    \label{tab:zero123_nvs_supp}
\end{table}

%% file: tex/5_conclusion.tex
\vspace{-1em}
\section{Conclusion}
\label{sec:conclusion}
We propose DreamComposer++, a flexible and scalable framework to empower existing diffusion models for zero-shot novel view synthesis with multi-view conditioning. DreamComposer++ is scalable to the number of input views. It can be flexibly plugged into a range of existing state-of-the-art models to empower them to generate high-fidelity novel view images or videos with multi-view conditions, ready for controllable 3D object reconstruction.

%% file: tex/acknowledgements.tex
\vspace{-1em}
\section*{Acknowledgements}
\label{sec:acknow}
This research was partially supported by the Research Grants Council of Hong Kong (T45-701/22-R).